\pdfoutput=1

\documentclass[11pt]{article}

\usepackage[dvipsnames]{xcolor}

\usepackage[]{acl}

\usepackage{times}
\usepackage{latexsym}

\usepackage[T1]{fontenc}

\usepackage[utf8]{inputenc}

\usepackage{microtype}

\usepackage{inconsolata}

\usepackage{amsmath,amsfonts,bm,mathtools,amssymb}
\usepackage{booktabs}
\usepackage{graphicx}
\usepackage{multirow}
\usepackage{nccmath}
\usepackage{makecell}
\usepackage{footmisc}
\usepackage{cleveref}
\usepackage{xcolor,colortbl}
\usepackage{subcaption}

\newcommand{\gcell}{\cellcolor[rgb]{0.8,0.95, 0.8}}
\newcommand{\rcell}{\cellcolor[rgb]{0.95, 0.8, 0.8}}

\usepackage{cleveref}
\crefname{section}{\S}{\S\S}
\crefformat{section}{\S#2#1#3}

\title{Psychometric Predictive Power of Large Language Models}

\author{Tatsuki Kuribayashi${}^{1}$ \qquad Yohei Oseki${}^{2}$ \qquad Timothy Baldwin${}^{1,3}$\\
        ${}^{1}$MBZUAI\quad
        ${}^{2}$The University of Tokyo \quad
        ${}^{3}$The University of Melbourne \\
  \texttt{\{tatsuki.kuribayashi,timothy.baldwin\}@mbzuai.ac.ae} \\
   \texttt{oseki@g.ecc.u-tokyo.ac.jp}}

\begin{document}
\maketitle
\begin{abstract}
Instruction tuning aligns the response of large language models (LLMs) with human preferences.
Despite such efforts in human--LLM alignment, we find that instruction tuning does not always make LLMs human-like from a cognitive modeling perspective. More specifically, next-word probabilities estimated by instruction-tuned LLMs are often worse at simulating human reading behavior than those estimated by base LLMs.
In addition, we explore prompting methodologies for simulating human reading behavior with LLMs. 
Our results show that prompts reflecting a particular linguistic hypothesis improve psychometric predictive power, but are still inferior to small base models.
These findings highlight that recent advancements in LLMs, i.e., instruction tuning and prompting, do not offer better estimates than direct probability measurements from base LLMs in cognitive modeling. In other words, pure next-word probability remains a strong predictor for human reading behavior, even in the age of LLMs.
\newline
\newline
\vspace{0.2em}
\hspace{.1em}\includegraphics[width=1.05em,height=1.05em]{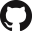}\hspace{.5em}\parbox{\dimexpr\linewidth-2\fboxsep-2\fboxrule}
  {\small \url{https://github.com/kuribayashi4/llm-cognitive-modeling}}
\end{abstract}

\section{Introduction}
\label{sec:intro}
\begin{figure}[t]
    \centering
    \includegraphics[width=7.5cm]{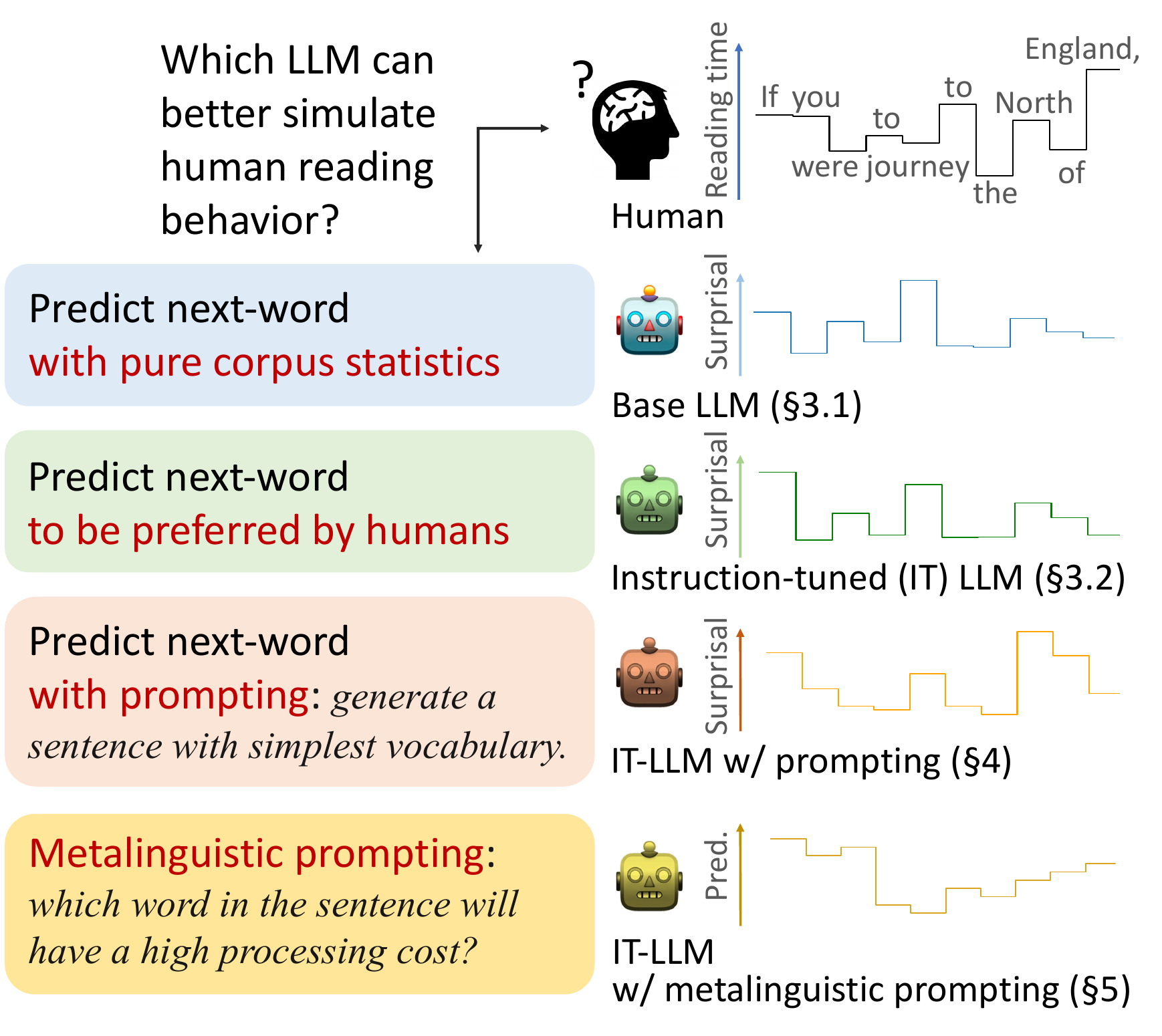}
    \caption{Comparing the ``reading behavior'' of humans and LLMs, i.e., reading time from humans is compared with surprisal from LLMs (\cref{subsec:linking}). We investigate which surprisal values estimated by: (i) base LLMs, (ii) instruction-tuned (IT) LLMs, (iii) IT-LLMs with prompting, or (iv) IT-LLMs with metalinguistic prompting can better simulate human reading time.}
    \label{fig:fig1}
\end{figure}

Aligning computational models with human perception/cognition has historically been a pivotal approach to understanding humans~\cite{Shapiro2003-pw}. With this in mind, computational psycholinguistics has investigated the model of human sentence processing~\cite{Crocker2010-cp} and recently found an intriguing correlation between next-word probabilities from language models (LMs) and human reading behavior---the less predictable a word is, the greater the cognitive load (e.g., longer reading time) humans exhibit---suggesting the expectation-based account of human sentence processing~\cite{Levy2008Expectation-basedComprehension,Smith2013TheLogarithmic}. Based on this finding, the field has further investigated which types of models/algorithms can compute probabilities better aligned with human reading behavior (Figure~\ref{fig:fig1};~\citet{hale-2001-probabilistic,Goodkind2018PredictiveQuality,Wilcox2020OnBehavior,Oh2021-ln,kuribayashi-etal-2022-context}; \textit{inter alia}).

In the field of natural language processing (NLP), in contrast, large language models (LLMs) tuned to human-preferred responses (e.g., GPT-3.5) improve in performance across a wide range of applications~\cite{ouyang2022training}.
Given the increasing prevalence of such human-aligned, \textit{instruction-tuned} LLMs (IT-LLMs), the following computational psycholinguistic question naturally arises: \textit{do IT-LLMs successfully simulate human reading behavior in terms of predicted surprisal?} 
The answer to this question is not immediately obvious. 
On the one hand, the answer might be \textit{yes} since these are tuned to human-preferred responses~\cite{zhang2023instruction}, which will be, broadly speaking, more aligned with human-like expectations of upcoming information during reading, e.g., not expect fabricated/hallucinated information during reading~\cite{grice1975logic,askell2021general}.
Moreover, some IT-LLMs employ reinforcement learning from human feedback (RLHF); such a scenario of language learning through (approximate) human feedback is more plausible than through text alone~\cite{alishahi2010computational}, and thus may enhance their cognitive plausibility.

On the other hand, IT-LLMs are a step beyond base LMs---pure statistical models of plausible text based on large natural language corpora---in the sense that IT-LLMs are tuned to specific human-preferred responses and may suffer from amplified reporting biases. 
If the core of human sentence processing is explained by next-word probabilities~\cite{Levy2008Expectation-basedComprehension,Smith2013TheLogarithmic}, instruction tuning will be unnecessary in simulating human reading behavior.
Based on this logic, the answer would be \textit{no}.

Our experiments show that IT-LLMs frequently yield worse psychometric predictive power (PPP) for human reading behavior than base LLMs (\cref{sec:exp}).
This suggests that the current paradigm of instruction tuning is irrelevant to modeling human sentence processing; IT-LLMs are not specially aligned with human language processing, at least based on reading time modeling, despite their ultimate goal of human--machine alignment.

Furthermore, we address open questions regarding prompting in cognitive modeling: (i) Can prompting result in probabilities from IT-LLMs being more aligned with human reading behavior? and (ii) Which is better at simulating human reading behavior, using direct probability measurements or directly asking IT-LLMs about the processing cost, e.g., \textit{metalinguistic prompting}~\cite{hu2023promptbased}?
For the first question, we find prompts in line with so-called ``good-enough'' human sentence processing~\cite{Ferreira2016-oj} to work well.
However, these are still worse than smaller base LLMs (\cref{sec:exp:prompt}).
For the second question, we find metalinguistic prompting to be inferior to direct probability measurement in terms of PPP (\cref{sec: metalinguistic}).

In sum, despite the recent advancements in IT-LLMs and prompting, they do not currently offer better estimates of human reading behavior than simple probability measurements from base LLMs.
This also underlines the value of access to probabilistic outputs for closed-source LLMs to further the study of cognitive modeling.

\section{Simulating human reading behavior}
\label{sec:modeling}

\subsection{Linking hypothesis}
\label{subsec:linking}
It has been reported that the word-by-word processing cost for humans, typically measured by reading time ($\mathrm{RT}$), can be explained by the surprisal of a word $h_{t,\theta}(w)$ in context $\bm w_{<t}=[w_0,\cdots,w_{t-1}]$, computed by a model $\theta$~\cite{hale-2001-probabilistic,Levy2008Expectation-basedComprehension,Smith2013TheLogarithmic}:
\begin{align}
    \label{eq:regression} 
    \mathrm{RT}(w_t) &\sim h_{t,\theta}(w_t) + \mathrm{baselines}(w_t) \;\;, \\
    \label{eq:surprisal} 
h_{t,\theta}(w) &:= -\log_2 p_\theta(w|\bm w_{<t}) \;\;.
\end{align}
To gauge the advantage of the surprisal factor in reading time modeling, we train two nested regression models\footnote{We used \texttt{statsmodels}~\cite{seabold2010statsmodels}.} (Eq.~\ref{eq:regression}) with and without the surprisal factor in addition to the $\mathrm{baselines}(w_t)$ factors.\footnote{We used the following formulation: \texttt{RT($w_t$) $\sim$ surprisal($w_t$) + surprisal($w_{t-1}$) + surprisal($w_{t-2}$) + length($w_t$) + freq($w_t$) + length($w_{t-1}$) + freq($w_{t-1}$) + length($w_{t-2}$) + freq($w_{t-2}$)}. The \texttt{surprisal($w_t$)} factor is excluded in the baseline regression model. The \texttt{freq($w_t$)} is quantified based on Wiki-103 data~\cite{Merity2016-rb} with logarithmic conversion. The \texttt{length($w_t$)} is the character length of the word. Adding an interaction term \texttt{length($w_i$)*freq($w_i$)} did not alter our findings, and thus we use the simpler independent model.}
Then, we report the psychometric predictive power (PPP), which is defined as the increase in the per-token average of the log-likelihood of the regression model due to the added surprisal factor.
A high PPP indicates the effectiveness of the surprisal factor in simulating human reading behavior.
Our interest in this paper is to find the model $\theta$ that leads to a higher PPP. 

Following existing studies~\cite{Roark2009-si,van-schijndel-linzen-2019-entropy,pimentel-etal-2022-effect}, we also examine other variants of Eq.~\ref{eq:regression} by replacing the surprisal factor $h_{t,\theta}(w)$ with the expected value of surprisal $\mathrm{H}_\theta(W_t)$, in the form of: (i) contextualized Shannon entropy~\cite{Shannon1948-jg}; and (ii) its generalization called contextualized R\'{e}nyi entropy $\mathrm{H}_{\alpha,\theta}(W_t)$~\cite{Renyi1961-sw}:

\begin{align}
    \mathrm{H}_\theta(W_t) &:= \underset{w\sim p(\cdot|\bm w_{<t})}{\mathbb{E}} h_{t,\theta}(w) \;\;, \\
    \mathrm{H}_{\alpha,\theta}(W_t) &:= \underset{\gamma\rightarrow\alpha}{\lim} \frac{1}{1-\gamma} \log_2 \sum_{w\in W} p_\theta(w|\bm w_{<t})^\gamma \;\;.
\end{align}
\noindent
Here, vocabulary set $W_t$ is approximated by the model's subword vocabulary set.\footnote{A set of entire words in natural language can inherently be infinite. See Appendix B in~\citet{pimentel-etal-2022-effect} for the details about subword-based entropy.}
For the R\'{e}nyi entropy, we set $\alpha=0.5$ based on the results of~\citet{pimentel-etal-2022-effect,Liu2023-ib}, noting that Shannon entropy is a special case of R\'{e}nyi entropy with $\alpha=1$.
Note that we limit $\bm w_{<t}$ to intra-sentential context since we are interested in sentence-level language processing in this study.
Word boundaries are based on the reading-time corpus; if a word consists of several subwords, cumulative surprisal is computed.

\subsection{Experimental settings}
\label{subsec:setup}

\paragraph{Models:}
We examined 26 LLMs as candidate models $\theta$ to compute information-theoretic values: six LLaMA-2~\cite{LLaMa}, four Falcon~\cite{falcon40b}, four GPT-2~\cite{Radford_undated-nn}, four GPT-3/3.5~\cite{ouyang2022training}\footnote{GPT-3 B2/D2 denotes \texttt{bebbage-002} and \texttt{davinci-002}, respectively.
GPT-3.5 D2/D3 denotes \texttt{text-davinci-002} and \texttt{text-davinci-003}, respectively.}, and eight OPT~\cite{opt} models with different sizes and instruction tuning settings (see Appendix~\ref{app:model} for details).
Among them, GPT-3.5, two LLaMA-2, and two Falcon models are IT-LLMs (models with \checkmark\ in the ``IT'' column in Table~\ref{tbl:ppp_ppl}), and the others are ``base LLMs.''
More specifically, GPT-3.5 D2 is trained through supervised fine-tuning, Falcon IT-LLMs are also trained via a particular supervised-tuning approach~\cite{xu-etal-2023-baize}\footnote{Further details are not clear at the time of writing this paper. We solely rely on the description here: \url{https://huggingface.co/tiiuae/falcon-40b-instruct}}, and GPT-3.5 D3 and LLaMA-2 IT-LLMs employ RLHF.
Note that entropy metrics are omitted from the GPT-3/3.5 results since their APIs do not provide the probability distribution over the entire vocabulary.

\paragraph{Data:}
We use two corpora: Dundee Corpus (DC:~\citet{kennedy2003dundee}) and  Natural Stories Corpus (NS:~\citet{futrell-etal-2018-natural}).
DC is eye-tracking data, where we use the first-pass duration as reading time, while NS is self-paced reading time data.
Following recent studies~\cite{Wilcox2020OnBehavior,Wilcox2021-gy,pimentel-etal-2022-effect}, we averaged the reading times for each word across different human subjects.
We excluded data points with a reading time of zero or beyond three standard deviations.
We also excluded the sentence-initial/final words since IT-LLMs tend to predict special phrases (e.g., \textit{Sure, here is my answer...}) at the sentence-initial position, and sentence-final words potentially have a confounding influence~\cite{Rayner2000TheReading,meister-etal-2022-analyzing}.

\begin{figure*}[t]
    \centering
    \includegraphics[width=\linewidth]{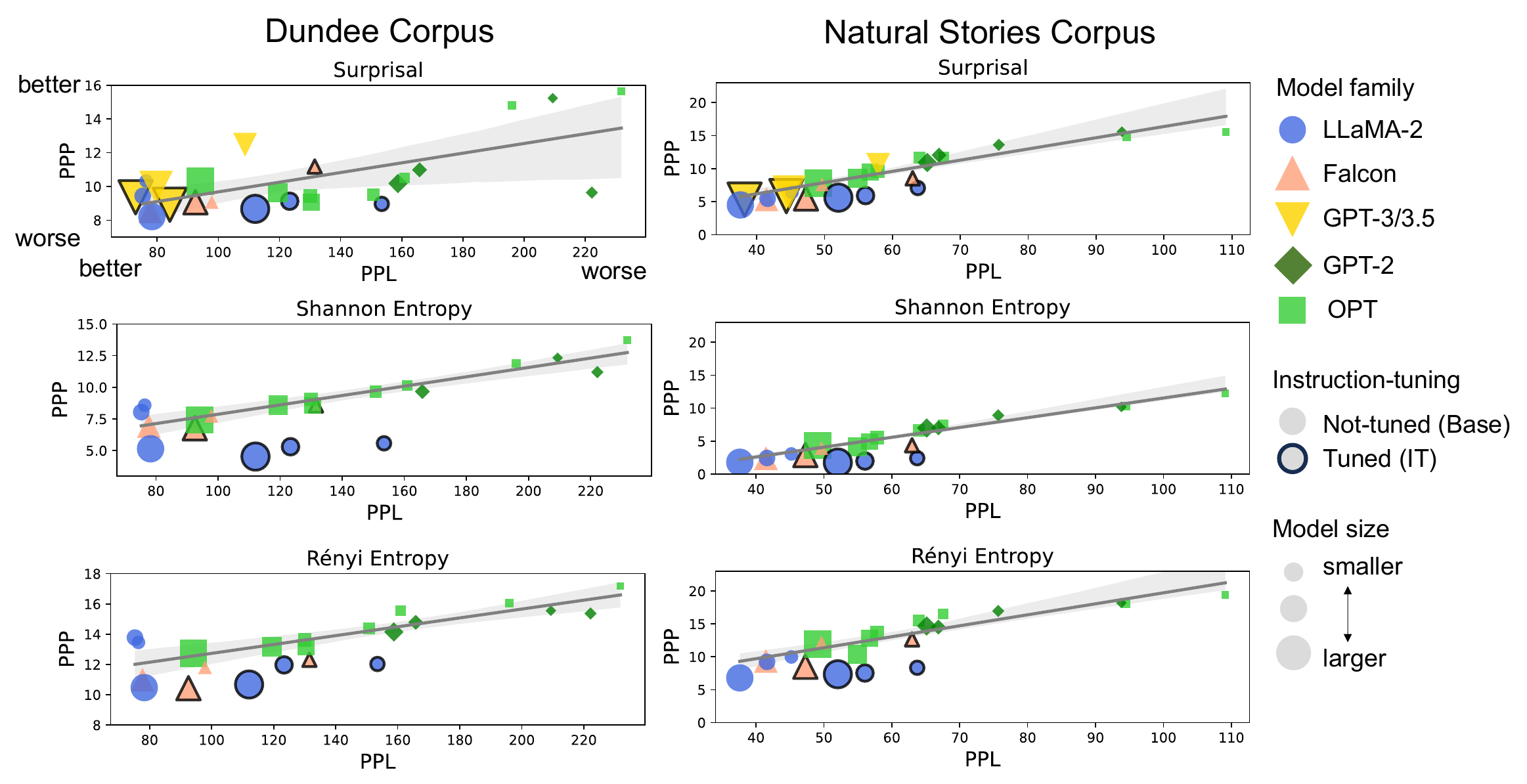}
    \caption{The relationship between PPL and PPP (see exact scores in Table~\ref{tbl:ppp_ppl}). Each point corresponds to each LLM, and those with a black edge line are IT-LLMs. The regression line is estimated by base LLMs, and the colored area presents a 95\% confidence interval.
    IT-LLMs were relatively poor (below the line) at balancing PPL and PPP.}
    \label{fig:ppl_ppp}
\end{figure*}

\begin{table}[t]
\centering
\fontsize{8}{9.5}\selectfont
\setlength{\tabcolsep}{0.2pt}
\begin{tabular}{lcrrrrrrrr}
\toprule
& & \multicolumn{4}{c}{DC} & \multicolumn{4}{c}{NS} \\
\cmidrule(l){3-6}  \cmidrule(l){7-10}
Model & IT & $h$ $\uparrow$ & $\mathrm{H}$ $\uparrow$ & $\mathrm{H_{0.5}}$ $\uparrow$ & PPL $\downarrow$ & $h$ $\uparrow$ & $\mathrm{H}$ $\uparrow$ & $\mathrm{H_{0.5}}$  $\uparrow$ & PPL $\downarrow$ \\
\cmidrule(r){1-1} \cmidrule(){2-2} \cmidrule(l){3-3}  \cmidrule(l){4-4} \cmidrule(l){5-5} \cmidrule(l){6-6}  \cmidrule(l){7-7} \cmidrule(l){8-8} \cmidrule(l){9-9} \cmidrule(l){10-10}
LMA-2 7B  &    &  10.33  &  8.58  &  13.45  &  76.40 	&  6.41  &  3.06  &  9.97  &  45.21 \\
LMA-2 7B  &  \checkmark  &  \rcell 8.97  &  \rcell 5.57  & \rcell 12.03  &  153.46 	 & \gcell 7.07  & \rcell 2.42  & \rcell 8.33  &  63.74 \\
LMA-2 13B  &    &  9.44  &  8.04  &  13.77  &  75.28 	 &  5.44  &  2.44  &  9.23  &  41.62 \\
LMA-2 13B  &  \checkmark  &  \rcell 9.13  &  \rcell 5.30  & \rcell 11.97  &  123.35 	 & \gcell 5.93  &  \rcell 1.99  & \rcell 7.53  &  56.05 \\
LMA-2 70B  &    &  8.21  &  5.14  &  10.47  &  78.28 	 &  4.51  &  1.80  &  6.79  &  37.61 \\
LMA-2 70B  &  \checkmark  & \gcell 8.67  & \rcell 4.53  & \gcell 10.67  &  112.07 	 & \gcell 5.60  & \rcell 1.75  & \gcell 7.34  &  52.05 \\
\cmidrule(r){1-1} \cmidrule(){2-2} \cmidrule(l){3-3}  \cmidrule(l){4-4} \cmidrule(l){5-5} \cmidrule(l){6-6}  \cmidrule(l){7-7} \cmidrule(l){8-8} \cmidrule(l){9-9} \cmidrule(l){10-10}
Falcon 7B  &    &  9.08  &  7.75  &  11.81  &  97.86 	 &  7.61  &  3.95  &  12.17  &  49.64 \\
Falcon 7B  &  \checkmark  & \gcell 11.18  & \gcell 8.57  & \gcell 12.31  &  131.53 	 & \gcell 8.54  & \gcell 4.38  &  \gcell 12.63  &  62.99 \\
Falcon 40B  &    &  8.53  &  6.93  &  10.99  &  77.72 	 &  5.35  &  2.41  &  9.36  &  41.46 \\
Falcon 40B  &  \checkmark  & \gcell 9.06  & \rcell 6.76  & \rcell 10.43  &  92.53 	 & \gcell 5.49  & \gcell 2.89  &  \rcell 8.49  & 47.27 \\
\cmidrule(r){1-1} \cmidrule(){2-2} \cmidrule(l){3-3}  \cmidrule(l){4-4} \cmidrule(l){5-5} \cmidrule(l){6-6}  \cmidrule(l){7-7} \cmidrule(l){8-8} \cmidrule(l){9-9} \cmidrule(l){10-10}
GPT-3 B2  &    &  12.47  &  -  &  -  &  108.77 	 &  10.58  &  -  &  -  &  57.91 \\
GPT-3 D2  &    &  9.93  &  -  &  -  &  79.65 	 &  6.45  &  -  &  -  &  44.79 \\
GPT-3.5 D2  &  \checkmark  & \rcell 9.35  &  -  &  -  &  72.95 	 & \rcell 5.30  &  -  &  -  &  38.23 \\
GPT-3.5 D3  &  \checkmark  & \rcell 8.91  &  -  &  -  &  84.17 	 & \rcell 5.83  &  -  &  -  &  44.38 \\
\cmidrule(r){1-1} \cmidrule(){2-2} \cmidrule(l){3-3}  \cmidrule(l){4-4} \cmidrule(l){5-5} \cmidrule(l){6-6}  \cmidrule(l){7-7} \cmidrule(l){8-8} \cmidrule(l){9-9} \cmidrule(l){10-10}
GPT-2 177M  &    &  15.23  &  12.32  &  15.55  &  209.37 	 &  15.61  &  10.20  &  18.19  &  93.81 \\
GPT-2 355M  &    &  9.63  &  11.20  &  15.37  &  222.17 	 &  13.62  &  8.91  &  16.96  &  75.67 \\
GPT-2 774M  &    &  10.98  &  9.66  &  14.79  &  165.81 	 &  12.04  &  7.01  &  14.52  &  66.87 \\
GPT-2 1.5B  &    &  10.18  &  -  &  14.15  &  158.75 	 &  10.94  &  6.99  &  14.69  &  65.14 \\
\cmidrule(r){1-1} \cmidrule(){2-2} \cmidrule(l){3-3}  \cmidrule(l){4-4} \cmidrule(l){5-5} \cmidrule(l){6-6}  \cmidrule(l){7-7} \cmidrule(l){8-8} \cmidrule(l){9-9} \cmidrule(l){10-10}
OPT 125M  &    &  15.65  &  13.72  &  17.18  &  231.80 	 &  15.54  &  12.27  &  19.41  &  109.11 \\
OPT 350M  &    &  14.81  &  11.89  &  16.07  &  196.02 	 &  14.86  &  10.35  &  18.11  &  94.51 \\
OPT 1.3B  &    &  10.51  &  10.16  &  15.55  &  160.95 	 &  11.81  &  7.43  &  16.53  &  67.59 \\
OPT 2.7B  &    &  9.52  &  9.65  &  14.38  &  150.78 	 &  11.66  &  6.60  &  15.51  &  63.98 \\
OPT 6.7B  &    &  9.43  &  9.06  &  13.63  &  130.01 	 &  9.59  &  5.56  &  13.64  &  57.86 \\
OPT 13B  &    &  9.06  &  8.57  &  13.15  &  130.44 	 &  9.51  &  4.96  &  12.84  &  56.74 \\
OPT 30B  &    &  9.62  &  8.58  &  13.17  &  119.42 	 &  8.55  &  4.16  &  10.39  &  54.91 \\
OPT 66B  &    &  10.30  &  7.42  &  12.73  &  94.15 	 &  7.78  &  4.33  &  11.92  &  49.11 \\
\bottomrule
\end{tabular}
\caption{The PPL and PPP scores of tested LMs. The ``IT'' column denotes whether instruction tuning is applied.  The columns $h$, $\mathrm{H}$, and $\mathrm{H_{0.5}}$ indicate surprisal, Shannon entropy, and R\'{e}nyi entropy ($\alpha=0.5$), respectively. The colors of cells for IT-LLMs indicate if the PPP \colorbox{green!15}{increased} or \colorbox{red!15}{decreased} relative to its base version (GPT-3.5 models are compared to GPT-3s).``LMA-2'' denotes the LLaMA-2 family.}
\label{tbl:ppp_ppl}
\end{table}

\section{Experiment 1: PPP of LLMs}
\label{sec:exp}
We first observe the PPP of base LLMs (\cref{subsec:replication}) and then analyze the PPP of IT-LLMs (\cref{subsec:instruction}). 
We explore prompting in \cref{sec:exp:prompt} and~\cref{sec: metalinguistic}.

\subsection{Reproducing previous findings}
\label{subsec:replication}
Table~\ref{tbl:ppp_ppl} shows the PPP and perplexity (PPL)\footnote{We measure the PPL of an LLM as average surprisal power of 2: $2^{\frac{1}{N} \sum_t h_{t,\theta}(w)}$ over the reading-time-annotated units; this ensures comparable PPL scores across LLMs with different tokenizers.} of each LLM. 
We first examine whether we are able to reproduce results from existing studies.

\paragraph{Surprisal theory:}
Across all the settings of \{model$\times$corpus$\times$metric\}, information-theoretic metrics are effective in simulating reading time as shown in existing studies~\cite[\textit{inter alia}]{Shain2022-qv}.
The PPP scores were significantly positive (F-test, $p<0.001$), and the coefficients of interest were also significantly larger than zero (one-sample two-sided $t$-test, $p<0.001$).

\paragraph{Advantage of R\'{e}nyi entropy:}
R\'{e}nyi entropy with $\alpha=0.5$ ($\mathrm{H_{0.5}}$) is consistently better at simulating human reading behavior (Table~\ref{tbl:ppp_ppl}) than the other metrics ($h$ and $\mathrm{H}$) in all settings.
Such an advantage has been reported with GPT-2 models~\cite{pimentel-etal-2022-effect,Liu2023-ib}; we further show that this generalizes to other LLMs.

\paragraph{PPP--PPL inverse scaling:}
While early work showed that better PPL leads to better PPP~\cite{Roark2009-si,frank2011insensitivity,Goodkind2018PredictiveQuality}, more recent work using neural LLMs has reported the opposite, inverse relationship between PPP and PPL when using base LLMs---the \textbf{worse} the PPL is, the better the PPP is~\cite{kuribayashi-etal-2022-context,Shain2022-qv,de-varda-marelli-2023-scaling,Oh2023-zw}, implying a ``superhuman'' ability of LLMs in next-word prediction.
The results of Table~\ref{tbl:ppp_ppl} are mapped onto the PPP and PPL axes in Figure~\ref{fig:ppl_ppp}.
Each point corresponds to an LLM, and those with black-edged outlines correspond to IT-LLMs (their results are analyzed in~\cref{subsec:instruction}).
The regression lines of the PPP--PPL relationship are estimated only by base LLMs; these lines replicate the inverse scaling effect, i.e., better PPL leads to worse PPP.
Pearson's correlation between PPP and PPL was positive in all settings ($r$ was within $0.55$--$0.95$ with $p<0.05$).
Our results confirm that such a relationship holds even when using LLaMA-2, Falcon, and GPT-3/3.5 models and entropy metrics, extending the results of earlier studies~\cite{Oh2023-zw}.

\subsection{The effects of instruction tuning}
\label{subsec:instruction}

\paragraph{Instruction tuning often hurts PPP:}
The green and red cells in Table~\ref{tbl:ppp_ppl} indicate the positive and negative effects of instruction tuning, respectively, based on comparison of the base model and its instruction-tuned variant (e.g., LLaMA-2 7B vs.\ LLaMA-2 7B instruct).
There are no consistent positive or negative effects and thus no evidence that instruction tuning causes LLMs to be more cognitively plausible in terms of cognitive modeling.
More specifically, LLaMA-2 and GPT-3.5 models tend to degrade under instruction tuning, while Falcon IT-LLMs show a somewhat positive effect.
Falcon family IT-LLMs employ a supervised-tuning approach~\cite{xu-etal-2023-baize} rather than RLHF, suggesting that RLHF might lead to drastic negative effects. 
Nevertheless, there could be several confounding factors, e.g., the base model architecture, training regimen, and instruction-tuning data/policies, motivating future work to investigate this effect in a more controlled manner.

\paragraph{Worse PPP compared to base LLMs with equivalent PPL:}
We additionally report a consistent tendency that IT-LLMs yielded poorer PPP than the base LLMs with equivalent PPL, on top of the reported PPP-PPL trade-off~\cite{Oh2023-zw}.
Figure~\ref{fig:ppl_ppp} shows that IT-LLMs (points with black-edged outlines) are positioned below the PPP--PPL regression line estimated by the base LLMs across all metrics and corpora.
Specifically, IT-LLMs yield poor PPL scores, presumably due to their objective no longer being pure language modeling.
Worse still, they also yield worse PPP scores than the expected good values based on the estimated inverse PPP-PPL relationship (gray lines in Figure~\ref{fig:ppl_ppp}).
Specifically, 32 results out of 34 \{IT-LLM$\times$metric$\times$corpus\} settings are below the regression line. This is significantly more frequent than chance $\pi=0.5$ based on a two-sided binomial test ($p<1\mathrm{e}-7$).
That is, IT-LLMs struggle to balance PPP and PPL.\footnote{To handle the concern of IT-LLMs simply being confused for a given sentence fragment $\bm w_{<t}$ in isolation, we re-conducted the experiments with the explicit instruction to predict the next word, yielding results consistent with the original; that is, the results hold up even after instructing IT-LLMs to behave as base LLMs (Appendix~\ref{app:ppp_result}).}

\paragraph{Discussion:}
These results indicate that instruction tuning \textbf{does not enhance the simulation of human reading behavior}, despite it being intended to make LLMs more human-aligned.
There are at least two hypotheses for why this should be the case: (i) instruction tuning with manipulated text amplifies reporting biases in training data and corrupts the LLM's language statistics built on naturally occurring text during pretraining; and (ii) the instruction-tuning objective is misaligned with human reading, e.g., IT-LLMs are trained to correctly predict responses over a huge range of questions, well beyond the capacity of a single human subject.
In other words, our results support the surprisal theory that links human sentence processing efforts with pure word surprisal~\cite{Levy2008Expectation-basedComprehension,Smith2013TheLogarithmic,Shain2022-qv}.

\section{Experiment 2: prompting}
\label{sec:exp:prompt}

\begin{table}[t!]
\centering
\fontsize{9}{9}\selectfont
\setlength{\tabcolsep}{3pt}
\begin{tabular}{lp{6cm}}
\toprule
ID & Prompt \\
\cmidrule(){1-1} \cmidrule(l){2-2}
Syn$\downarrow$ & Please complete the following sentence to make it as grammatically simple as possible: \texttt{\textbackslash n} $w_0, \cdots, w_{t-1}$ \\
\cmidrule(){1-1} \cmidrule(l){2-2}
Syn$\rightarrow$ & Please complete the following sentence with a careful focus on grammar: \texttt{\textbackslash n} $w_0, \cdots, w_{t-1}$ \\
\cmidrule(){1-1} \cmidrule(l){2-2}
Syn$\uparrow$ & Please complete the following sentence to make it as grammatically complex as possible: \texttt{\textbackslash n} $w_0, \cdots, w_{t-1}$ \\
\cmidrule(){1-1} \cmidrule(l){2-2}
Lex$\downarrow$ & Please complete the following sentence using the simplest vocabulary possible: \texttt{\textbackslash n} $w_0, \cdots, w_{t-1}$ \\
\cmidrule(){1-1} \cmidrule(l){2-2}
Lex$\rightarrow$ & Please complete the following sentence with a careful focus on word choice: \texttt{\textbackslash n} $w_0, \cdots, w_{t-1}$ \\
\cmidrule(){1-1} \cmidrule(l){2-2}
Lex$\uparrow$ & Please complete the following sentence using the most difficult vocabulary possible: \texttt{\textbackslash n} $w_0, \cdots, w_{t-1}$ \\
\cmidrule(){1-1} \cmidrule(l){2-2}
Task1 & Please complete the following sentence in a human-like manner. It has been reported that human ability to predict next words is weaker than language models and that humans often make noisy predictions, such as careless grammatical errors. \texttt{\textbackslash n} $w_0, \cdots, w_{t-1}$ \\
\cmidrule(){1-1} \cmidrule(l){2-2}
Task2 & Please complete the following sentence. We are trying to reproduce human reading times with the word prediction probabilities you calculate, so please predict the next word like a human. It has been reported that human ability to predict next words is weaker than language models and that humans often make noisy predictions, such as careless grammatical errors. \texttt{\textbackslash n} $w_0, \cdots, w_{t-1}$ \\
\cmidrule(){1-1} \cmidrule(l){2-2}
Base & Please complete the following sentence: \texttt{\textbackslash n} $w_0, \cdots, w_{t-1}$ \\
\bottomrule
\end{tabular}
\caption{Our examined prompts. The IDs shown in the first column are also used in Tables~\ref{tbl:sent_stats} and~\ref{tbl:prompt_result}.}
\label{tbl:prompts}
\end{table}

The use of IT-LLMs stimulates an additional question---\textit{does prompt tuning lead to better PPP for IT-LLMs?}
One can control the \textit{prior belief} of LLMs about upcoming words through the prompt, e.g., \textit{complete the sentence to make it grammatically simple.}
Analyzing effective prompt types may shed light on underlying bias in human expectation-based reading, i.e., which kinds of words are more expected by humans.

\begin{table}[t]
\centering
\fontsize{8}{9}\selectfont
\setlength{\tabcolsep}{1.5pt}
\begin{subtable}[h]{0.45\textwidth}
        \centering
\begin{tabular}{lrrrrrrrrrr}
\toprule
& \multicolumn{5}{c}{dependency length $\uparrow$} & \multicolumn{5}{c}{sentence length $\uparrow$} \\ 
\cmidrule(l){2-6} \cmidrule(l){7-11}
& \multicolumn{3}{c}{LLaMA-2} & \multicolumn{2}{c}{Falcon} & \multicolumn{3}{c}{LLaMA-2} & \multicolumn{2}{c}{Falcon} \\ 
\cmidrule(l){2-4} \cmidrule(l){5-6} \cmidrule(l){7-9}  \cmidrule(l){10-11}
Prompt & 7B & 13B & 70B & 7B & 40B & 7B & 13B & 70B & 7B & 40B \\ 
\cmidrule(){1-1} \cmidrule(l){2-6} \cmidrule(l){7-11}
Syn$\downarrow$ & 2.10 & 2.43 & 2.51 & 2.27 & 2.48 & 12.3 & 14.6 & 16.4 & 12.8 & 15.7 \\
Syn$\rightarrow$ & 2.90 & 2.87 & 2.68 & 2.65 & 3.47 & 24.9 & 22.9 & 21.2 & 18.4 & 18.1 \\
Syn$\uparrow$ & \textbf{3.45} & \textbf{3.29} & \textbf{3.31} & 2.86 & \textbf{3.49} & \textbf{43.9} & \textbf{44.0} & \textbf{45.5} & \textbf{23.8} & \textbf{33.5} \\
\cmidrule(){1-1} \cmidrule(l){2-4} \cmidrule(l){5-6} \cmidrule(l){7-9}  \cmidrule(l){10-11}
Lex$\downarrow$ & 2.31 & 2.40 & 2.46 & 2.28 & 3.01 & 13.4 & 13.1 & 14.7 & 13.5 & 15.7 \\
Lex$\rightarrow$ & 2.95 & 3.29 & 3.05 & 2.58 & 3.33 & 32.4 & 28.0 & 25.2 & 17.5 & 19.9 \\
Lex$\uparrow$ & 3.08 & 3.06 & 3.24 & 2.86 & 3.23 & 35.2 & 36.5 & 33.8 & 24.6 & 27.6 \\
\cmidrule(){1-1} \cmidrule(l){2-4} \cmidrule(l){5-6} \cmidrule(l){7-9}  \cmidrule(l){10-11}
Task1 & 2.87 & 2.94 & 2.82 & 2.70 & 2.87 & 27.1 & 28.2 & 28.6 & 21.1 & 26.2 \\
Task2 & 2.99 & 2.80 & 3.07 & 2.79 & 2.82 & 23.6 & 20.4 & 18.3 & 21.2 & 21.4 \\
\cmidrule(){1-1} \cmidrule(l){2-4} \cmidrule(l){5-6} \cmidrule(l){7-9}  \cmidrule(l){10-11}
Base & 2.65 & 2.47 & 2.77 & \textbf{2.91} & 3.00 & 17.9 & 17.4 & 19.3 & 20.4 & 18.9 \\
\bottomrule
\end{tabular}
 \caption{Syntactic complexity}
       \label{tbl:syn_comp}
\end{subtable}
\\
\vspace{0.3cm}
\begin{subtable}[h]{0.45\textwidth}
        \centering
\begin{tabular}{lrrrrrrrrrr}
\toprule
& \multicolumn{5}{c}{$\log$ word frequency $\downarrow$} & \multicolumn{5}{c}{word length $\uparrow$} \\ 
\cmidrule(l){2-6} \cmidrule(l){7-11}
& \multicolumn{3}{c}{LLaMA-2} & \multicolumn{2}{c}{Falcon} & \multicolumn{3}{c}{LLaMA-2} & \multicolumn{2}{c}{Falcon} \\ 
\cmidrule(l){2-4} \cmidrule(l){5-6} \cmidrule(l){7-9}  \cmidrule(l){10-11}
Prompt & 7B & 13B & 70B & 7B & 40B & 7B & 13B & 70B & 7B & 40B \\ 
\cmidrule(){1-1} \cmidrule(l){2-6} \cmidrule(l){7-11}
Syn$\downarrow$ & 4.77 & 4.89 & 4.69 & 4.87 & 4.87 & 3.97 & 4.08 & 4.23 & 3.78 & 4.13 \\
Syn$\rightarrow$ & 4.55 & 4.67 & 4.63 & 4.79 & 4.84 & 4.44 & 4.53 & 4.46 & 3.94 & 4.39 \\
Syn$\uparrow$ & 4.43 & 4.44 & 4.55 & 4.64 & 4.70 & 4.60 & 4.68 & 4.67 & 4.84 & 4.35 \\
\cmidrule(){1-1} \cmidrule(l){2-4} \cmidrule(l){5-6} \cmidrule(l){7-9}  \cmidrule(l){10-11}
Lex$\downarrow$ & 4.73 & 4.80 & 4.73 & 4.79 & 4.85 & 4.16 & 3.62 & 4.22 & 4.03 & 4.11 \\
Lex$\rightarrow$ & 4.55 & 4.76 & 4.57 & 4.50 & 4.84 & 4.49 & 4.51 & 4.54 & 4.59 & 4.31 \\
Lex$\uparrow$ & \textbf{3.90} & \textbf{4.09} & \textbf{3.80} & \textbf{4.37} & \textbf{4.44} & \textbf{5.16} & \textbf{4.98} & \textbf{5.27} & \textbf{4.99} & \textbf{4.95} \\
\cmidrule(){1-1} \cmidrule(l){2-4} \cmidrule(l){5-6} \cmidrule(l){7-9}  \cmidrule(l){10-11}
Task1 & 4.75 & 4.73 & 4.68 & 4.65 & 4.78 & 4.61 & 4.31 & 4.47 & 4.94 & 4.27 \\
Task2 & 4.80 & 4.75 & 4.83 & 4.87 & 4.80 & 4.28 & 4.14 & 4.23 & 4.37 & 4.46 \\
\cmidrule(){1-1} \cmidrule(l){2-4} \cmidrule(l){5-6} \cmidrule(l){7-9}  \cmidrule(l){10-11}
Base & 4.77 & 4.86 & 4.87 & 4.93 & 4.91 & 4.37 & 4.21 & 4.23 & 4.15 & 4.16 \\
\bottomrule
\end{tabular}
 \caption{Lexical complexity. Stopwords are excluded when calculating the average of logarithmic word frequencies.}
       \label{tbl:lex_comp}
\end{subtable}
\caption{Statistics of sentences generated with different prompts and IT-LLMs. The highest values of dependency, sentence, and word length, and the lowest value of $\log$ word frequency for each model are highlighted.}
\label{tbl:sent_stats}
\end{table}

\begin{table*}[t]
\centering
\fontsize{8}{8}\selectfont
\setlength{\tabcolsep}{3pt}
\begin{tabular}{lrrrrrrrrrrrrrr}
\toprule
& \multicolumn{7}{c}{DC} & \multicolumn{7}{c}{NS} \\
 \cmidrule(l){2-8} \cmidrule(l){9-15}
& \multicolumn{3}{c}{$h$ $\uparrow$} & \multicolumn{2}{c}{$\mathrm{H}$ $\uparrow$} & \multicolumn{2}{c}{$\mathrm{H_{0.5}}$ $\uparrow$} & \multicolumn{3}{c}{$h$ $\uparrow$} & \multicolumn{2}{c}{$\mathrm{H}$ $\uparrow$} & \multicolumn{2}{c}{$\mathrm{H_{0.5}}$ $\uparrow$} \\
\cmidrule(l){2-4} \cmidrule(l){5-6} \cmidrule(l){7-8} \cmidrule(l){9-11} \cmidrule(l){12-13} \cmidrule(l){14-15}
Prompt & LMA-2 & Falcon & GPT-3.5 & LMA-2 & Falcon & LMA-2 & Falcon & LMA-2 & Falcon & GPT-3.5 & LMA-2 & Falcon & LMA-2 & Falcon \\
\cmidrule(){1-1} \cmidrule(l){2-2} \cmidrule(l){3-3} \cmidrule(l){4-4} \cmidrule(l){5-5} \cmidrule(l){6-6} \cmidrule(l){7-7} \cmidrule(){8-8} \cmidrule(l){9-9} \cmidrule(l){10-10} \cmidrule(l){11-11} \cmidrule(l){12-12} \cmidrule(l){13-13} \cmidrule(l){14-14} \cmidrule(l){15-15}
Syn$\downarrow$ & 8.20 & \textbf{9.73} & 6.20 & 5.94 & \textbf{9.77} & 12.30 & \textbf{12.69} & 3.84 & 7.38 & \textbf{5.83} & 2.57 & 4.60 & 8.78 & 13.30 \\
Syn$\rightarrow$ & 8.63 & 9.46 & 6.56 & 5.56 & 9.52 & 11.94 & 12.40 & 4.34 & \textbf{7.46} & 5.23 & 1.88 & 4.63 & 7.14 & 13.01 \\
Syn$\uparrow$ & 8.18 & 9.46 & 5.90 & 5.64 & 9.60 & 11.66 & 12.56 & 4.64 & 7.10 & 4.07 & 1.31 & 4.09 & 6.38 & 12.35 \\
\cmidrule(){1-1} \cmidrule(l){2-4} \cmidrule(l){5-6} \cmidrule(l){7-8} \cmidrule(l){9-11} \cmidrule(l){12-13} \cmidrule(l){14-15}
Lex$\downarrow$ & 7.93 & 9.49 & 5.92 & 6.33 & 9.74 & 12.39 & 12.63 & 3.61 & 6.77 & 4.79 & 2.00 & 4.74 & 8.34 & 13.26 \\
Lex$\rightarrow$ & 8.10 & 9.16 & \textbf{6.63} & 5.18 & 9.18 & 11.34 & 12.28 & 3.79 & 7.30 & 5.30 & 1.37 & 4.21 & 6.36 & 12.75 \\
Lex$\uparrow$  & 7.79 & 8.81 & 5.77 & 5.34 & 9.29 & 11.44 & 12.34 & 3.66 & 6.01 & 3.88 & 1.12 & 3.65 & 6.00 & 11.53 \\
\cmidrule(){1-1} \cmidrule(l){2-4} \cmidrule(l){5-6} \cmidrule(l){7-8} \cmidrule(l){9-11} \cmidrule(l){12-13} \cmidrule(l){14-15}
Task1 & 8.83 & 8.82 & 5.54 & 5.99 & 8.86 & 12.39 & 12.35 & 4.29 & 5.46 & 2.95 & 2.61 & 4.73 & \textbf{9.64} & 13.42 \\
Task2 & \textbf{8.90} & 9.16 & 5.95 & \textbf{6.53} & 9.13 & \textbf{13.11} & 12.42 & \textbf{4.76} & 5.99 & 3.23 & \textbf{2.64} & \textbf{5.14} & 9.43 & \textbf{13.54} \\
\cmidrule(){1-1} \cmidrule(l){2-4} \cmidrule(l){5-6} \cmidrule(l){7-8} \cmidrule(l){9-11} \cmidrule(l){12-13} \cmidrule(l){14-15}
Base & 8.42 & 9.60 & 6.98 & 5.65 & 9.45 & 12.15 & 12.41 & 4.94 & 7.13 & 4.69 & 1.94 & 5.01 & 7.82 & 13.52 \\
\cmidrule(){1-1} \cmidrule(l){2-4} \cmidrule(l){5-6} \cmidrule(l){7-8} \cmidrule(l){9-11} \cmidrule(l){12-13} \cmidrule(l){14-15}
W/o & 8.92 & 10.12 & 9.13 & 5.13 & 7.67 & 11.56 & 11.37 & 6.20 & 7.02 & 5.56 & 2.05 & 3.63 & 7.73 & 10.56 \\
\bottomrule
\end{tabular}

\caption{The PPP scores when using each prompt (the highest scores other than baseline ones for each corpus/metric are in boldface). Scores are averaged in each IT-LLM family (``LMA-2'' denotes LLaMA-2). The columns $h$, $\mathrm{H}$, and $\mathrm{H_{0.5}}$ indicate surprisal, Shannon entropy, and R\'{e}nyi entropy ($\alpha=0.5$) settings, respectively. ``W/o'' denotes the setting without any prompt.}
\label{tbl:prompt_result}
\end{table*}

\paragraph{Settings:}
We examine the nine prompts shown in Table~\ref{tbl:prompts} to linguistically bias the LLM-computed information-theoretic values.\footnote{We used a slightly different prompting format for LLaMA-2s (see Appendix~\ref{app:prompt_list}).}
The first six prompts focus on syntactic and lexical complexity, based on longstanding interest in syntactic and lexical biases in expectation-based reading~\cite{Roark2009-si,frank2011insensitivity}.
The ``Task1'' and ``Task2'' prompts inform IT-LLMs of the task-specific objective of our experiments.
A prompt is appended immediately before the context words ($\bm w_{<t}$ in Eq.~\ref{eq:surprisal}) when computing the information-theoretic values.
That is, we now use \textbf{prompt-conditioned surprisal, Shannon entropy, and R\'{e}nyi entropy} ($\alpha=0.5$) values with a given prompt $r$:

\begin{align}
    &h_{t,\theta}(w, r) := -\log_2 p_\theta(w|\bm w_{<t}, r)  \;\;, \\
    &\mathrm{H}_\theta(W_t, r) := \underset{w\sim p(\cdot|\bm w_{<t})}{\mathbb{E}} h_{t,\theta}(w, r) \;\;, \\
    \nonumber
    &\mathrm{H}_{\alpha,\theta}(W_t, r) \\
    &:= \underset{\gamma\rightarrow\alpha}{\lim} \frac{1}{1-\gamma} \log_2\sum_{w\in W} p_\theta(w|\bm w_{<t}, r)^\gamma \;\;.
\end{align}

\subsection{Preliminary: does prompting control next-word prediction?}
\label{subsec:prompt_ling_stats}

To ensure that prompting does indeed induce the intended changes in the next-word distribution, we first analyze sentences generated with different prompts using LLaMA-2 and Falcon IT-LLMs.
To diversify the input context, we first extract 20 sentences from the DC dataset and append their first five words to each of the nine prompts, resulting in 180 (=20$\times$9) input contexts.
Then, we generate 180 sentences by feeding the respective input contexts to the IT-LLMs.
To measure the syntactic and lexical biases in text generation, we report the averaged dependency length, sentence length, logarithmic word frequency, and word length of the sentences generated with each prompt.
Appendix~\ref{app:prompt_stats} provides more details of this preliminary analysis.

\paragraph{Prompting controls next-word prediction as intended:}
Table~\ref{tbl:sent_stats} shows the statistics of sentences generated by different prompts.
As intended, syntactically complex, long sentences with long syntactic dependencies are generated when the model is instructed to make sentences grammatically complex (Syn$\uparrow$; Table~\ref{tbl:syn_comp}), and vice versa (Syn$\downarrow$; Table~\ref{tbl:syn_comp}).
The lexical biases are also appropriately injected by prompting: the Lex$\uparrow$ (Lex$\downarrow$) prompts result in sentences with less (more) frequent and longer (shorter) words (Table~\ref{tbl:lex_comp}).
On the basis of this, we can infer that LLMs have some ability to bias sentence completion prediction based on linguistic instructions. 
This finding itself opens up the potential of prompting as a way of controlling the linguistic bias in information-theoretic values rather than, e.g., training differently-biased LLMs from scratch as typically done in computational psycholinguistics research~\cite{frank2011insensitivity}.
Note that, nevertheless, LLMs are not able to perfectly separate the two orthogonal dimensions of syntactic and lexical complexity in text generation with prompts. For example, Lex$\downarrow$ and Lex$\uparrow$ prompts somewhat affect the syntactic complexity of generated sentences (Table~\ref{tbl:syn_comp}) as well.
Progress in controlled text generation should mitigate this effect~\cite{pmlr-v202-zhou23g}.

\begin{figure*}[t]
    \centering
    \includegraphics[width=\linewidth]{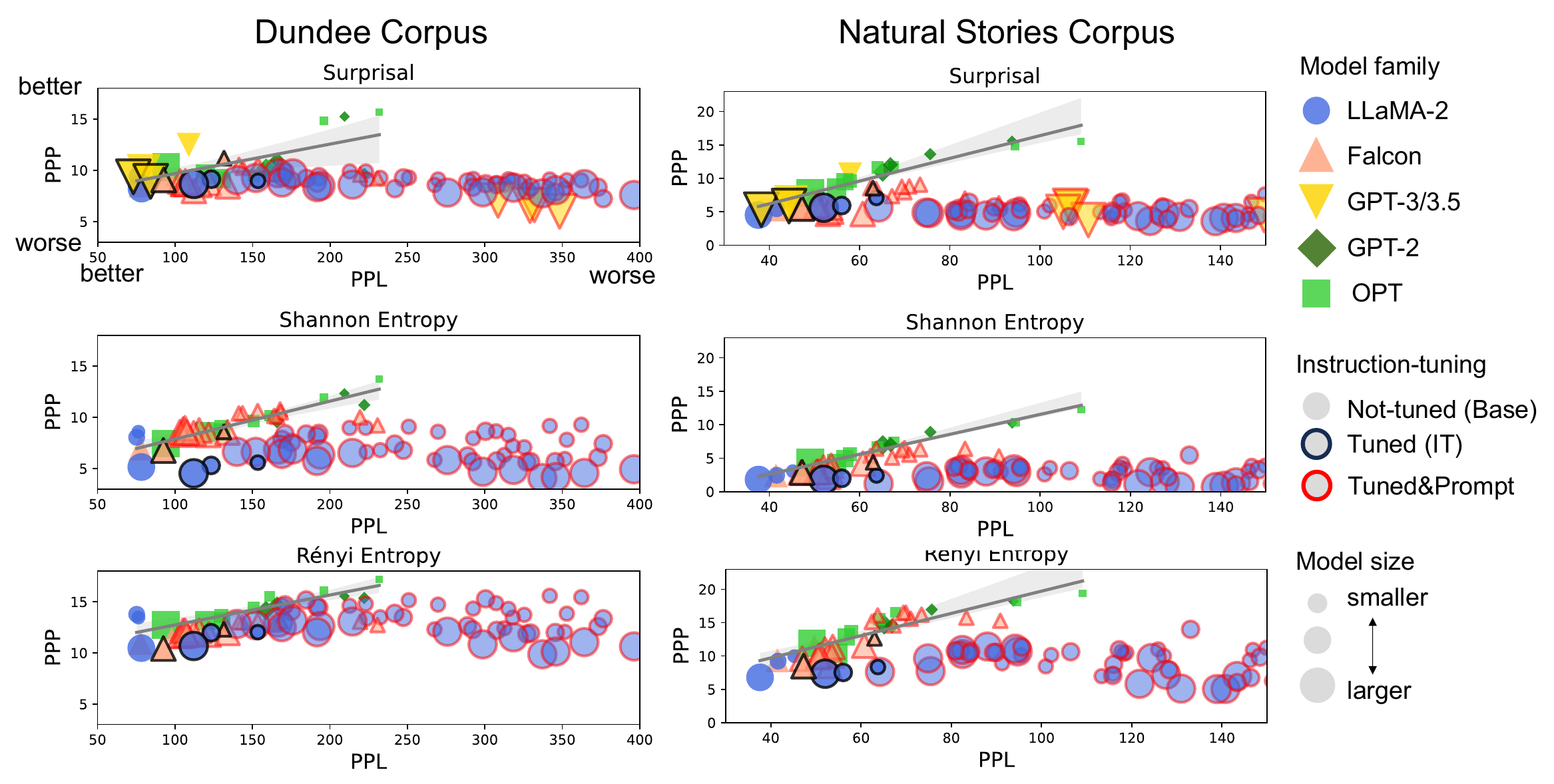}
    \caption{The PPL and PPP of LLMs with prompting are plotted at the top of Figure~\ref{fig:ppl_ppp}. Each point corresponds to a given combination of LLM and prompt, and those with red-edged outlines are IT-LLMs with a particular prompt. The PPP--PPL regression line is estimated by base LLMs, and the colored area presents the 95\% confidence interval.
    IT-LLMs with prompting are poorer than base LLMs at balancing PPL and PPP.}
    \label{fig:ppl_ppp_prompt}
\end{figure*}

\subsection{Results: PPP of prompt-conditioned LLMs}
\paragraph{Effective prompts:}
Table~\ref{tbl:prompt_result} shows the averaged PPP against each prompt in each IT-LLM family: instruction-tuned LLaMA-2 (7B, 13B, 70B), Falcon (7B, 40B), and GPT-3.5 (D2 and D3) models.
We also report two baseline results obtained with the ``Base'' prompt (Table~\ref{tbl:prompts}) and without prompting.
We observe the following: (i) some prompts, typically, Syn$\downarrow$ and Task2, lead LLMs to achieve better PPP than baselines; (ii) such effective prompting strategies are generally consistent across corpora; and (iii) instructions to use simple grammar/vocabulary work better than those to use complex grammar/vocabulary.
In terms of the first finding, some prompt-conditioned LLMs exhibit better PPP than the baselines, especially based on the entropy measurements ($\mathrm{H}$ and $\mathrm{H_{0.5}}$).
For the second finding, particular prompts, e.g., Syn$\downarrow$ and Task2, generally result in the best PPP on both the DC and NS datasets.
Finally, for the third finding, prompts to use simple grammar/vocabulary (Syn$\downarrow$ and Lex$\downarrow$) lead to better PPP than the opposite instructions (Syn$\uparrow$ and Lex$\uparrow$).
Note that, interestingly, the task-specific instructions also generally worked well, especially for the entropy measures.
Detailed results of each model and prompt are given in Appendix~\ref{app:prompt_results}.

\paragraph{Discussion:}
Regarding the third finding of the advantage of prompts to use simple grammar/vocabulary, one plausible implication is that LLMs tend to predict more complex, wordy text completions than human expectations during reading, and this is alleviated somewhat by these prompts.
Such an implied \textbf{simplicity bias} in effective prompts potentially lends support to the theory of so-called good-enough processing in human language processing~\cite{Ferreira2016-oj}.

\subsection{Analysis: PPP--PPL relationship}
\paragraph{Prompt-conditioned LLMs underperform base LLMs with equivalent PPLs:}
Figure~\ref{fig:ppl_ppp_prompt} shows the PPP and PPL of IT-LLMs conditioned on different prompts, including those listed in Appendix~\ref{app:prompt_list}, superimposed on the results from Figure~\ref{fig:ppl_ppp}.
Similar to the results in~\cref{sec:exp}, the prompt-conditioned LLMs (points with red-edged outlines) are under the PPP--PPL regression line estimated from base LLMs; 448 results out of 468 \{prompt$\times$model$\times$metric$\times$corpus\} settings are below the line.
This is significantly more frequent than chance $\pi=0.5$ based on a two-sided binomial test ($p<1\mathrm{e}-105$).
This shows that the base LLMs set the empirical Pareto front with respect to the PPP--PPL trade-off against prompt-biased LLMs, and in other words, none of the prompt-conditioned IT-LLMs could outperform the small base LMs with the best PPP, such as GPT-2 small.
That is, base LLMs are a strong baseline in reading time modeling.

\section{Experiment 3: metalinguistic prompting}
\label{sec: metalinguistic}
Instead of using direct probability measurements, one can also ask IT-LLMs about the processing cost of words via \textit{metalinguistic} prompting~\cite{hu2023promptbased}, e.g., \textit{Please estimate the cognitive load of this word in context}.
We explore such an approach in this section.

\paragraph{Settings:}
Based on preliminary experiments, asking a model to predict ther eading time for each word via prompting does not work well.
Thus, we make the problem simpler: rank words in a sentence in order of processing difficulty---\textit{which word in this sentence incurs a higher cognitive load during reading? Please order the words (high to low cost).}
We ask such a question to the model, then calculate Spearman's rank correlation between the estimated word processing costs and their actual reading time in each sentence.\footnote{If the LLMs output a word not in the target sentence, or the output lacks particular words in the targeted sentence, these tokens are excluded in computing the correlation.} 
Then, these correlation scores are averaged across the sentences in the reading time corpus; a high score indicates the model's estimated being more aligned with reading time.\footnote{100 and 50 sentences are used for the DC and NS experiments, respectively. These are the first five sentences in each document in the respective corpus (DC/NS have 20/10 documents). We partially observed that these limited-scale experiments can approximate the full-scale experiments.}
We employed a 3-shot setting,\footnote{Three exemplars are from the opposing corpus: the NS sentences were used for the DC experiments, and vice versa.} and ran three different runs with different exemplars for prompting.
The average and standard deviation scores across the three runs are reported (Table~\ref{tbl:prompt_estimate}).
We also calculated the rank correlation between surprisal and reading time as a baseline.
LLaMA-2 70B is excluded due to computation resource limitations.

\subsection{Results}
\paragraph{Direct probability measurements outperform metalinguistic prompting:}
Table~\ref{tbl:prompt_estimate} shows the examined prompts and results (see Appendix~\ref{app:prompt_examples} for the exact prompts).
The standard surprisal-based method yields higher correlations than metalinguistic prompting methods, which yield near-zero correlations.
Specifically, correlations from surprisal-based methods were significantly larger than those from metalinguistic prompting (two-sided Mann-Whitney U test, $p<1\mathrm{e}-8$) in both corpora.
We suspect that the model simply failed in \textit{ordering many items}; thus, we calculated the rank correlation coefficient by only using the first five words listed by the model, but the correlations were again nearly zero.
Such inferiority is consistent with the results in grammaticality judgment tasks~\cite{hu2023promptbased}.

\begin{table}[t]
\centering
\fontsize{8}{9}\selectfont
\setlength{\tabcolsep}{0.5pt}
\begin{tabular}{p{3cm}lrr}
\toprule

 Prompt/method & Model & DC $\uparrow$ & NS $\uparrow$ \\
\cmidrule(){1-1} \cmidrule(l){2-2} \cmidrule(l){3-3} \cmidrule(l){4-4} 
 \multirow{6}{3cm}{\makecell[l]{Suppose humans read the \\ following sentence: \\ (SENT). List the tokens in \\ order of their reading cost \\ (high to low) during \\ sentence processing.}} & LMA-2 7B & 0.09$\pm$0.02  & -0.04$\pm$0.06  \\
 & LMA-2 13B & 0.06$\pm$0.02 & -0.03$\pm$0.06  \\
 & Falcon 7B & 0.12$\pm$0.01  & 0.01$\pm$0.09  \\
  & Falcon 40B & 0.03$\pm$0.04 & $-$0.03$\pm$0.11  \\
  & GPT3.5 D2 & 0.05$\pm$0.03 & 0.05$\pm$0.03  \\
  & GPT3.5 D3 & 0.08$\pm$0.03 & 0.03$\pm$0.02 \\
 \cmidrule(){1-1} \cmidrule(l){2-2} \cmidrule(l){3-3} \cmidrule(l){4-4} 
  \multirow{6}{3cm}{\makecell[l]{Suppose you read the \\ following sentence: \\ (SENT). 
List the tokens in \\ order of their probability \\ in context (low to high).}} & LMA-2 7B & 0.05$\pm$0.06 & 0.00$\pm$0.02  \\
 & LMA-2 13B & 0.04$\pm$0.03 & 0.06$\pm$0.04  \\
 & Falcon 7B & 0.08$\pm$0.05 & 0.05$\pm$0.02   \\
  & Falcon 40B  & 0.02$\pm$0.07 & 0.13$\pm$0.10  \\
   & GPT3.5 D2 & 0.03$\pm$0.00 & 0.02$\pm$0.00   \\
  & GPT3.5 D3 & $-$0.01$\pm$0.02  & 0.06$\pm$0.03  \\
  \cmidrule(){1-1} \cmidrule(l){2-2} \cmidrule(l){3-3} \cmidrule(l){4-4} 
  \multirow{6}{3cm}{Surprisal-based \\ estimation} & LMA-2 7B & 0.28 & 0.19   \\
 & LMA-2 13B & 0.27 & 0.19  \\
 & Falcon 7B & 0.32 & 0.18  \\
   & Falcon 40B & 0.28 & 0.17  \\
   & GPT3.5 D2 & 0.28 & 0.16  \\
  & GPT3.5 D3 & 0.25 & 0.17 \\
\bottomrule
\end{tabular}
\caption{Average and standard deviation of rank correlation $\rho$ between estimated cognitive load and reading time of words across three different runs. ``LMA-2'' denotes the LLaMA-2 family.}
\label{tbl:prompt_estimate}
\end{table}

\begin{table}[t]
\centering
\fontsize{8}{9}\selectfont
\setlength{\tabcolsep}{3pt}
\begin{tabular}{lrr}
\toprule
 Model & DC $\uparrow$ & NS $\uparrow$ \\
\cmidrule(){1-1}  \cmidrule(l){2-2} \cmidrule(l){3-3}
LLaMA-2 7B  & 0.12$\pm$0.13& 0.15$\pm$0.08 \\
LLaMA-2 13B  & 0.02$\pm$0.10& 0.06$\pm$0.07 \\
Falcon 7B & 0.15$\pm$0.08 & 0.30$\pm$0.09  \\
Falcon 40B & 0.09$\pm$0.09 & 0.17$\pm$0.00  \\
GPT3.5 D2 & 0.15$\pm$0.02 & 0.22$\pm$0.07  \\
GPT3.5 D3 & 0.18$\pm$0.05 & 0.24$\pm$0.02 \\
\bottomrule
\end{tabular}
\caption{Rank correlation $\rho$ between the word probability (rank) estimated by the prompt and the actual surprisal values computed by the corresponding model.}
\label{tbl:meta-cognition}
\end{table}

\subsection{Analysis: metacognition of own surprisal}
\paragraph{Gap between computed probability and response to metalinguistic prompts:}
\textit{Are the estimates of word probability based on metalinguistic prompting consistent with their actual surprisal?}
To answer this question, we measure Spearman's rank correlation between the probabilities of words estimated via metalinguistic prompting (the second prompt in Table~\ref{tbl:prompt_estimate}) and their actual surprisal values.
The correlations are around 0.1--0.2 (Table~\ref{tbl:meta-cognition}), suggesting that metalinguistic prompting about word probability is again not an accurate measure of actual surprisal.

\section{Related work}
\paragraph{Simulating human reading behavior:}
Expectation-based accounts of human reading have been actively explored based on the linking hypothesis between surprisal and human reading behavior~\cite{Levy2008Expectation-basedComprehension,Smith2013TheLogarithmic}.
To gain insights into black-box human sentence processing, analysis has been done on which types of models/algorithms better simulate human reading behavior~\cite{hale-2001-probabilistic,Demberg2008-fd,frank2011insensitivity,Goodkind2018PredictiveQuality,Aurnhammer2019-fu,Wilcox2020OnBehavior,Merkx2020ComparingData,kuribayashi-etal-2021-lower,noji-oseki-2021-effective,Oh2021-ln,Michaelov2021-cx}.
It has been reported that those with specific properties, e.g., syntactic operations~\cite{Hale2018FindingSearch,Yoshida2021-rc}, memory limitations~\cite{kuribayashi-etal-2022-context,Timkey2023-dy}, and/or appropriate input units~\cite{Oh2021-ln,nair2023words} yield better fit to human reading behavior.
Building on this body of work, we show that the current generation of IT-LLMs offers a somewhat poor predictor in cognitive modeling.

\paragraph{Prompt-based analysis of linguistic knowledge in LLMs:}
Given the rise of the prompting paradigm, testing the linguistic knowledge of LLMs via prompting has gained recent attention~\cite{Li2022-hm,hu2023promptbased,Katzir2023,Begus2023,Dentella2023-jg,blevins-etal-2023-prompting}.
Prior work has pointed out their inferior ability at linguistic judgments under metalinguistic prompting to directly estimate probabilities~\cite{hu2023promptbased,Dentella2023-jg}, and specifically~\citet{hu2023promptbased} dubbed this discrepancy the so-called competence--performance distinction~\cite{chomsky1965aspects} of LLMs.
This problem is also related to the calibration of model outputs~\cite{kadavath2022language}.
We revealed such degradation of metalinguistic prompting methods in simulating human reading behavior (\cref{sec: metalinguistic}).

\paragraph{Instruction tuning:}
Starting from the multitask fine-tuning of LMs~\cite{wei2021finetuned,sanh2022multitask}, \textit{instruction-tuning}---aligning a model with human users---has played a crucial role in developing LLMs~\cite{ouyang2022training,glaese2022improving}. 
The objective of instruction tuning is, for example,  making models \textit{helpful}, \textit{honest}, and \textit{harmless}~\cite{askell2021general} in addition to just following the instructions.
Notably, researchers may have been aligning LLMs to not the exact model of humans but rather a superhuman chat agent with instruction tuning; our results might reflect the paradox---pursuing human preferences has resulted in creating something different from humans.

Concurrent with this study, others have investigated the effects of instruction-tuning on the cognitive plausibility and linguistic knowledge of LLMs~\cite{Aw2023-tx,Kauf2024-lx}.
In particular, \citet{Aw2023-tx} suggest that instruction tuning leads to a divergence in brain alignment and behavioral alignment, consistent with our work.
The connection with our work requires further exploration.

\section{Conclusion}
We investigated the PPP of instruction-tuned LLMs, given their popularity in NLP.
We found that IT-LLMs yield worse PPP than base LLMs with equivalent perplexity, demonstrating the ineffectiveness of current instruction tuning and (metalinguistic) prompting in simulating human reading behavior.
One important area of future work is to explore why the current instruction-tuning paradigm is ineffective for modeling human reading behavior, and we highlighted the direction as \textit{aligning LLMs with the human cognition/perception}, which has historically been a scientific approach to understanding humans, a.k.a.\ cognitive modeling, as well as with practical chat agents.

\section*{Limitations}
Revealing why instruction tuning leads to a degradation in PPP is an important open question (\cref{subsec:instruction}).
In particular, ablating reinforcement learning from human feedback (RLHF) via controlled experiments would be an interesting research direction.
Unfortunately, the exact resources used in the instruction tuning of GPTs, LLaMA-2, and Falcon are not available, making it difficult to ablate the instruction-tuning scenario.
Instead, training and evaluating separate LMs with different instruction tuning scenarios using publicly-available resources for instruction tuning~\cite{DatabricksBlog2023DollyV2,alpaca} will be needed to further investigate our observation.

The scale of our experiments was limited across at least three dimensions.
First, we only targeted the English language, although, to some degree, some language-dependent observations related to reading time have been obtained by prior work~\cite{Vasishth2010-ji,Frank2016-so,kuribayashi-etal-2021-lower,siegelman2022expanding,Wilcox2023-pi,pouw-etal-2023-cross}.
Second, we tested only three families of IT-LLMs.
Specifically, at the time of finalizing this paper, the GPT-3.5 models were no longer accessible, and newer versions do not offer the option to output the probability of the generated text; this motivates a focus on open LLMs for this line of research.
Third, the variety of examined prompts was somewhat limited. 
Scaling up experiments with respect to these points is an obvious area for future work.
Note that the use of pre-trained LMs poses the possibility of data leakage of source texts. \citet{wilcox-etal-2023-language} suggest that such leakage may not be a primary factor in characterizing PPP; thus, we tentatively put this concern aside in this study.

\section*{Ethical considerations}
We do \textbf{not} in any way claim that the superiority of base LLMs in cognitive modeling, which potentially have harmful biases which are mitigated by instruction tuning, entails that human language processing also has inherently harmful biases.
All we have shown in practice is a general macro-trend that surprisal from base LLMs has better PPP than instruction-tuned models, and such a potential bias in human language processing should be carefully inspected in separate work, e.g., as per~\citet{Lior2023-jb}.
We used some writing assistance tools, e.g., ChatGPT and Grammarly, in the writing of this paper, just to fix language errors.

\section*{Acknowledgement}
We would like to thank Yova Kementchedjhieva, Go Inoue, and Ted Briscoe for their insightful feedback on earlier drafts of this paper. 
This work was supported in part by JST PRESTO Grant Number JPMJPR21C2, Japan.

\bibliography{custom}

\clearpage
\appendix

\section{Models}
\label{app:model}
We used the LLM implementations available via the huggingface transformer library~\cite{wolf-etal-2020-transformers}.
The exact model URLs are listed in Table~\ref{tbl:model}.
Some large models are loaded using quantization.
We used them for text decoding or probability computation; such usage of models follows their license and intended use.
A single NVIDIA A100 GPU (40GB) was used for the experiments.

As of the date we experimented (2023/10/20), we can not get the probability information from recent OpenAI models such as GPT-4~\cite{gpt4}. Thus, we only used legacy models that can yield probability scores. We also excluded the \texttt{gpt-3.5-turbo-instruct} model since the use of logprobs and echo options is restricted for this model to compute probabilities.

\begin{table*}[t]
    \centering
    \fontsize{8}{10}\selectfont
\setlength{\tabcolsep}{3pt}
    \begin{tabular}{lclr}
\toprule
Model & Instruction-tuning & Link & Quant. \\
\cmidrule(){1-1} \cmidrule(l){2-2} \cmidrule(l){3-3} \cmidrule(l){4-4}
GPT-2 117M & & \url{https://huggingface.co/gpt2} & \\
GPT-2 355M & & \url{https://huggingface.co/gpt2-medium} & \\
GPT-2 774M & & \url{https://huggingface.co/gpt2-large} & \\
GPT-2 1.5B & & \url{https://huggingface.co/gpt2-xl} & \\
\cmidrule(){1-1} \cmidrule(l){2-2} \cmidrule(l){3-3} \cmidrule(l){4-4}
LLaMa2 7B & & \url{https://huggingface.co/meta-llama/Llama 2-7b-hf} & \\
LLaMa2 7B & \checkmark & \url{https://huggingface.co/meta-llama/Llama 2-7b-chat-hf} & \\
LLaMa2 13B & & \url{https://huggingface.co/meta-llama/Llama 2-13b-hf} & 8bits \\
LLaMa2 13B & \checkmark & \url{https://huggingface.co/meta-llama/Llama 2-13b-chat-hf} & 8bits \\
LLaMa2 70B &  & \url{https://huggingface.co/meta-llama/Llama 2-70b-hf}  & 4bits \\
LLaMa2 70B & \checkmark & \url{https://huggingface.co/meta-llama/Llama 2-70b-chat-hf}& 4bits \\
\cmidrule(){1-1} \cmidrule(l){2-2} \cmidrule(l){3-3} \cmidrule(l){4-4}
Falcon 7B &   & \url{ttps://huggingface.co/tiiuae/falcon-7b} & \\
Falcon 7B & \checkmark & \url{https://huggingface.co/tiiuae/falcon-7b-instruct} & \\
Falcon 40B &   & \url{https://huggingface.co/tiiuae/falcon-40b} & 4bits\\
Falcon 40B & \checkmark & \url{https://huggingface.co/tiiuae/falcon-40b-instruct}& 4bits \\
\cmidrule(){1-1} \cmidrule(l){2-2} \cmidrule(l){3-3} \cmidrule(l){4-4}
OPT 125M &   & \url{https://huggingface.co/facebook/opt-125m} &\\
OPT 350M &   & \url{https://huggingface.co/facebook/opt-350m} &\\
OPT 1.3B&   & \url{https://huggingface.co/facebook/opt-1.3b} &\\
OPT 2.7B &   & \url{https://huggingface.co/facebook/opt-2.7b} &\\
OPT 6.7B &   & \url{https://huggingface.co/facebook/opt-6.7b} &\\
OPT 13B &   & \url{https://huggingface.co/facebook/opt-13b} &\\
OPT 30B &   & \url{https://huggingface.co/facebook/opt-30b} &\\
OPT 66B &   & \url{https://huggingface.co/facebook/opt-66b} &\\
\cmidrule(){1-1} \cmidrule(l){2-2} \cmidrule(l){3-3} \cmidrule(l){4-4}
GPT-3 babbage-002 &   & accessed on 2023/10/20 for~\cref{sec:exp}, and on 2023/11/04 for~\cref{sec:exp:prompt} and~\cref{sec: metalinguistic} &  \\
GPT-3 davinci-002 &   & accessed on 2023/10/20 for~\cref{sec:exp}, and on 2023/11/04 for~\cref{sec:exp:prompt} and~\cref{sec: metalinguistic}  & \\
GPT-3.5 text-davinci-003 & \checkmark  & accessed on 2023/10/20 for~\cref{sec:exp}, and on 2023/11/04 for~\cref{sec:exp:prompt} and~\cref{sec: metalinguistic} &\\
GPT-3.5 text-davinci-002 & \checkmark  & accessed on 2023/10/20 for~\cref{sec:exp}, and on 2023/11/04 for~\cref{sec:exp:prompt} and~\cref{sec: metalinguistic}  & \\

\bottomrule
\end{tabular}
    \caption{Information about the LLMs.}
    \label{tbl:model}
\end{table*}

\section{PPL--PPP relationship with the explicit instruction to complete the sentence}
\label{app:ppp_result}
\begin{figure*}[t]
    \centering
    \includegraphics[width=15cm]{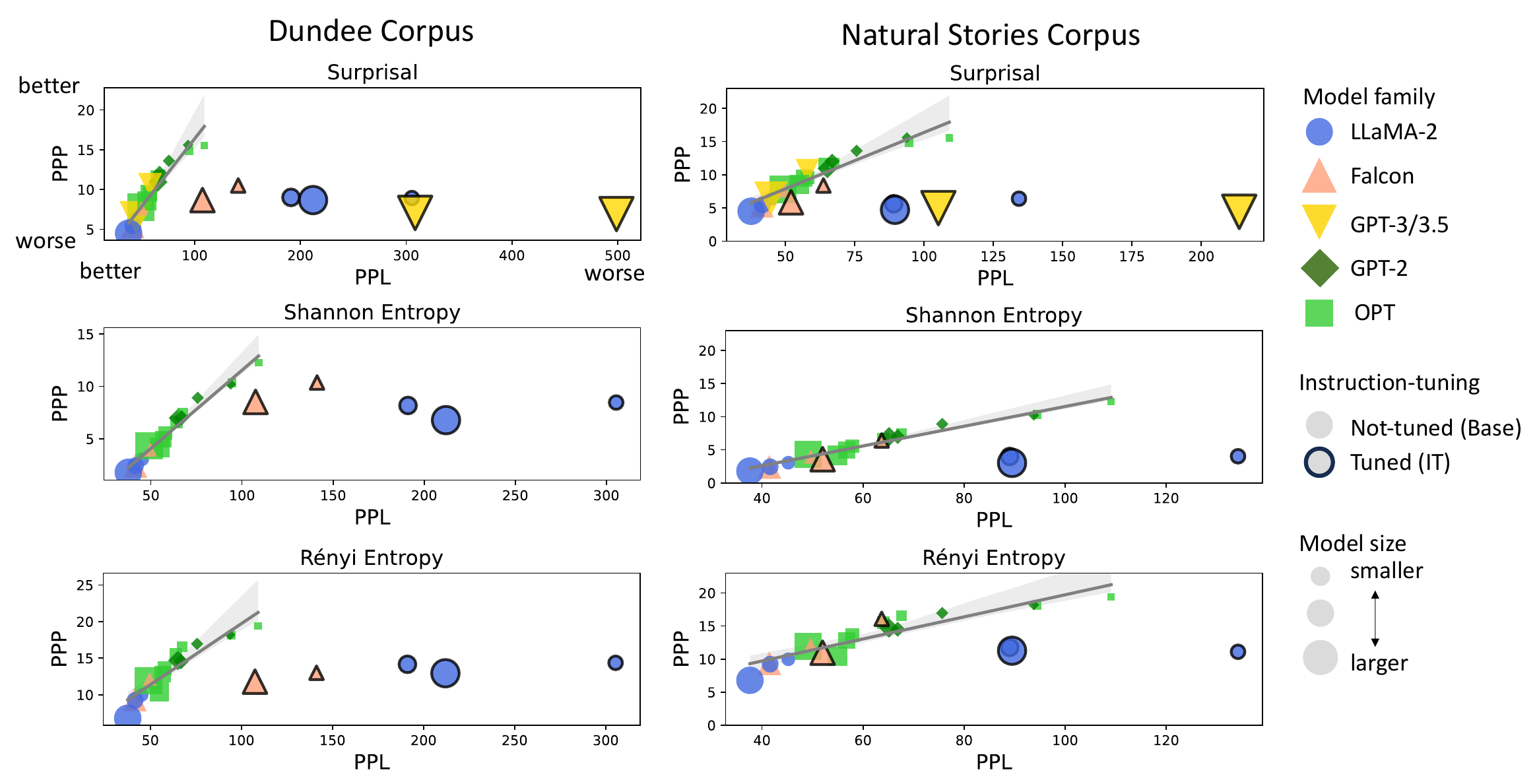}
    \caption{The relationship between PPL and PPP when using the prompt: \textit{Please complete the following sentence:}. Each point corresponds to each LLM, and those with black-edged outlines are IT-LLMs. The PPP--PPL regression line is estimated by base LLMs, and the colored area presents a 95\% confidence interval.
    IT-LLMs were relatively poor (below the line) at balancing PPL and PPP.}
    \label{fig:ppl_ppp_complete_only}
\end{figure*}

Figure~\ref{fig:ppl_ppp_complete_only} shows the same experiments as \cref{subsec:instruction} except for using the prompt: \textit{Please complete the following sentence:}.
The results were consistent with \cref{subsec:instruction}; the IT-LLMs exhibited competitive or worse PPP against the PPP--PPL relationship estimated by base LMs.

\section{Prompting}

\subsection{Prompt lists}
\label{app:prompt_list}
Table~\ref{tbl:prompt_list} shows the prompts used in our experiments.
We used the format-1 prompts for Falcon and GPT-3.5 models and the format-2 for LLaMA-2 models.
That is, the format-2 is used in~\cref{subsec:prompt_ling_stats}.
Table~\ref{tbl:prompt_result} in~\cref{sec:exp} shows the averaged results using the respective formats.
Figure~\ref{fig:ppl_ppp_prompt} shows all the results, including LLaMA-2 with format-1 prompts and Falcon and GPT-3.5 with format-2 prompts.

\paragraph{Why?}
We observed that LLaMA-2 models tend to generate additional user inquiries when using the format-1 prompts; thus, by format-2 prompts, we made it clear for LLaMA-2 models to play the system roles and special instruction marking symbols [INST].\footnote{Based on the LLaMA-2 format information: \url{https://github.com/samrawal/llama2_chat_templater}}
For example, LLaMA-2 completed the format-1 prompt (red words are generated) as follows: \textit{Please complete the following sentence: \textbackslash n They were playing on the \textcolor{red}{\_\_\_\_\_\_\_\_\_ when they found the treasure. \textbackslash n A. beach \textbackslash n B. mountain \textbackslash n C. park \textbackslash n D. forest \textbackslash n Answer: A. beach.}}
Conversely, with the format-1 prompt, Falcon and GPT-3.5 models are shown to complete the sentence intendedly and exhibit better PPLs than format-2 results; thus, we used the format-1 prompts for these LLMs.

\begin{table*}[t]
\centering
\fontsize{8}{8}\selectfont
\setlength{\tabcolsep}{1.5pt}
\begin{tabular}{p{7.5cm}p{7.5cm}}
\toprule
Format 1 (GPT3.5, Falcon) & Format 2 (LLaMA-2) \\
\cmidrule(){1-1} \cmidrule(l){2-2}
Please complete the following sentence to make it as grammatically simple as possible: \texttt{\textbackslash n} $w_0, \cdots, w_{t-1}$  & [INST] Please generate a grammatically simple sentence as much as possible. [/INST] Answer:  $w_0, \cdots, w_{t-1}$ \\
\cmidrule(){1-1} \cmidrule(l){2-2}
 Please complete the following sentence with a careful focus on grammar \texttt{\textbackslash n} $w_0, \cdots, w_{t-1}$ &  [INST] Please generate a sentence with a careful focus on grammar. [/INST]  Answer:  $w_0, \cdots, w_{t-1}$ \\
 \cmidrule(){1-1} \cmidrule(l){2-2}
 Please complete the following sentence to make it as grammatically complex as possible: \texttt{\textbackslash n} $w_0, \cdots, w_{t-1}$ & [INST] Please generate a grammatically complex sentence as much as possible. [/INST]  Answer: $w_0, \cdots, w_{t-1}$ \\
 \midrule
  Please complete the following sentence using the simplest vocabulary possible: \texttt{\textbackslash n} $w_0, \cdots, w_{t-1}$ & [INST] Please generate a sentence using the simplest vocabulary possible. [/INST] Answer:  \texttt{\textbackslash n} $w_0, \cdots, w_{t-1}$ \\
   \cmidrule(){1-1} \cmidrule(l){2-2}
   Please complete the following sentence with a careful focus on word choice \texttt{\textbackslash n} $w_0, \cdots, w_{t-1}$ & [INST] Please generate a sentence with a careful focus on word choice. [/INST] Answer:  \texttt{\textbackslash n} $w_0, \cdots, w_{t-1}$ \\
 \cmidrule(){1-1} \cmidrule(l){2-2}
 Please complete the following sentence using the most difficult vocabulary possible: \texttt{\textbackslash n} $w_0, \cdots, w_{t-1}$ & [INST] Please generate a sentence using the most difficult vocabulary possible. [/INST] Answer:  \texttt{\textbackslash n} $w_0, \cdots, w_{t-1}$ \\
 \midrule
 Please complete the following sentence in a human-like manner. It has been reported that human ability to predict next words is weaker than language models and that humans often make noisy predictions, such as careless grammatical errors. \texttt{\textbackslash n} $w_0, \cdots, w_{t-1}$ & [INST] Please generate a sentence in a human-like manner. It has been reported that human ability to predict next words is weaker than language models and that humans often make noisy predictions, such as careless grammatical errors. [/INST] Answer:  \texttt{\textbackslash n} $w_0, \cdots, w_{t-1}$ \\
 \cmidrule(){1-1} \cmidrule(l){2-2}
 Please complete the following sentence. We are trying to reproduce human reading times with the word prediction probabilities you calculate, so please predict the next word like a human. It has been reported that human ability to predict next words is weaker than language models and that humans often make noisy predictions, such as careless grammatical errors. \texttt{\textbackslash n} $w_0, \cdots, w_{t-1}$ & [INST] Please generate a sentence. We are trying to reproduce human reading times with the word prediction probabilities you calculate, so please predict the next word like a human. It has been reported that human ability to predict next words is weaker than language models and that humans often make noisy predictions, such as careless grammatical errors. [/INST] Answer:  \texttt{\textbackslash n} $w_0, \cdots, w_{t-1}$ \\
 \cmidrule(){1-1} \cmidrule(l){2-2}
 [INST] Please complete the following sentence: \texttt{\textbackslash n} $w_0, \cdots, w_{t-1}$ & Please generate a sentence. [/INST] Answer: \texttt{\textbackslash n} $w_0, \cdots, w_{t-1}$ \\
\bottomrule
\end{tabular}
\caption{All the prompts we used in our experiments.}
\label{tbl:prompt_list}
\end{table*}

\subsection{Details on prompt biases}
\label{app:prompt_stats}
\paragraph{Setting details:}
We appended the first five words in the second sentence in DC's each document to a particular prompt for sampling sentences.
When a model generated multiple sentences, the first one was used for the analysis in~\cref{subsec:prompt_ling_stats}.
We set top\_p threshold to 0.95.
In the analysis in~\cref{subsec:prompt_ling_stats}, Zipf frequency was computed with the \texttt{word\_freq} package~\cite{robyn_speer_2022_7199437}; here, stopwords were excluded using nltk.
Sentence and dependency lengths were computed using the nltk sentence/word tokenizers and spaCy dependency parser (\texttt{en\_core\_web\_sm} version). 
Notably, some of the statistics (dependency length distribution) form non-normal distribution; reporting the average score (Table~\ref{tbl:sent_stats}) would not be suitable.
Nevertheless, other statistics, such as skewness, yielded similar results.
Thus, we tentatively adopted the report of average values for simplicity.
Tables~\ref{tbl:prompt_example_generation3} and~\ref{tbl:prompt_example_generation} show the example of completed sentences conditioned by different prompts.

\begin{table*}[t]
    \centering
    \fontsize{8}{9.5}\selectfont
     \setlength{\tabcolsep}{1.5pt}
\begin{tabular}{p{15cm}}
\toprule

Please generate a grammatically simple sentence as much as possible.\\
Answer:\\
At that time he said, \textcolor{red}{"I am going to the store."}\\
\midrule
Please generate a sentence with a careful focus on grammar.\\
Answer:\\
At that time he said \textcolor{red}{he was going to the store, but he went to the park instead.}\\
\midrule
Please generate a grammatically complex sentence as much as possible.\\
Answer:\\
At that time he said \textcolor{red}{that he had seen her not more than once or twice since they had last met, when he had been so struck by her beauty that he had almost lost his breath, and he had not been able to forget her ever since, despite his many attempts}\\
\midrule
Please generate a sentence using the simplest vocabulary possible.\\
Answer:\\
At that time he said, \textcolor{red}{"I can't go."}\\
\midrule
Please generate a sentence with a careful focus on word choice.\\
Answer:\\
At that time he said \textcolor{red}{to himself, "I'm going to go out and do something that will make me feel really good."}\\
\midrule
Please generate a sentence using the most difficult vocabulary possible.\\
Answer:\\
At that time he said \textcolor{red}{to himself, "I must needs make an inscrutable and enigmatic allusion to the pervasive and pernicious effects of the miasmic and noxious atmosphere which permeates and pervades this putative}\\
\midrule
Please generate a sentence in a human-like manner. It has been reported that human ability to predict next words is weaker than language models and that humans often make noisy predictions, such as careless grammatical errors.\\
Answer:\\
At that time he said \textcolor{red}{he would be able to predict the next words but he was wrong.}\\
\midrule
Please generate a sentence. We are trying to reproduce human reading times with the word prediction probabilities you calculate, so please predict the next word like a human. It has been reported that human ability to predict next words is weaker than language models and that humans often make noisy predictions, such as careless grammatical errors.\\
Answer:\\
At that time he said \textcolor{red}{he was going to the store, but he actually went to the library.}\\
\midrule
Please generate a sentence.\\
Answer:\\
At that time he said \textcolor{red}{to me, "My dear friend, I must go to the store."}\\
\bottomrule
\end{tabular}
    \caption{Examples of the completions by LLaMA-2 70B with different prompts.}
    \label{tbl:prompt_example_generation3}
\end{table*}

\begin{table*}[t]
    \centering
    \fontsize{8}{9.5}\selectfont
     \setlength{\tabcolsep}{1.5pt}

    \begin{tabular}{p{15cm}}
    \toprule
    Please generate a grammatically simple sentence as much as possible.\\
Answer:\\
The evidence of the annual \textcolor{red}{rings is the most important proof that the tree grows.}\\
\midrule
    Please generate a sentence with a careful focus on grammar.\\
Answer:\\
The evidence of the annual \textcolor{red}{festival was apparent in the colorful decorations adorning the streets.}\\
\midrule
Please generate a grammatically complex sentence as much as possible.\\
Answer:\\
The evidence of the annual \textcolor{red}{festival's increasing popularity and cultural significance, as demonstrated by the growing number of attendees and the expanding variety of musical performances, food vendors, and artisanal goods on display, has made it imperative for the organiz}\\
\midrule
Please generate a sentence using the simplest vocabulary possible.\\
Answer:\\
The evidence of the annual \textcolor{red}{flooding was obvious.}\\
\midrule
Please generate a sentence with a careful focus on word choice.\\
Answer:\\
The evidence of the annual \textcolor{red}{festival was scattered across the grounds, with discarded flyers and abandoned souvenirs littering the once-pristine lawns.}\\
\midrule
Please generate a sentence using the most difficult vocabulary possible.\\
Answer:\\
The evidence of the annual \textcolor{red}{financial audit conducted by the external accounting firm indicated that the company's fiscal year-end consolidated balance sheet contained several glaring discrepancies, which necessitated a comprehensive reconciliation process to rectify the}\\
\midrule
Please generate a sentence in a human-like manner. It has been reported that human ability to predict next words is weaker than language models and that humans often make noisy predictions, such as careless grammatical errors.\\
Answer:\\
The evidence of the annual \textcolor{red}{performance review was scattered all over the table, making it difficult for the manager to piece together a coherent picture of the employee's strengths and weaknesses.}\\
\midrule
Please generate a sentence. We are trying to reproduce human reading times with the word prediction probabilities you calculate, so please predict the next word like a human. It has been reported that human ability to predict next words is weaker than language models and that humans often make noisy predictions, such as careless grammatical errors.\\
Answer:\\
The evidence of the annual \textcolor{red}{performance review was overwhelming.}\\
\midrule
Please generate a sentence.\\
Answer:\\
The evidence of the annual \textcolor{red}{rings in the cross section of a tree trunk provides valuable information to dendrochronologists.}\\
\bottomrule
\end{tabular}
    \caption{Examples of the completions by LLaMA-2 70B with different prompts.}
    \label{tbl:prompt_example_generation}
\end{table*}

\subsection{Detailed results}
\label{app:prompt_results}
Tables~\ref{tbl:prompt_surprisal_dc},~\ref{tbl:prompt_surprisal_dc2},~\ref{tbl:prompt_surprisal_ns}, and~\ref{tbl:prompt_surprisal_ns2} present the detailed results shown in \cref{sec:exp:prompt}.
The advantage of prompts to use simple vocabulary/grammar demonstrated in \cref{sec:exp:prompt} is generally reproduced across different models and corpora, but GPT 3.5 models sometimes prefer the instruction to ``carefully focus on grammar'' rather than use simple grammar.

\begin{table*}[t]
\centering
\fontsize{8}{8}\selectfont
\setlength{\tabcolsep}{1.5pt}
\begin{tabular}{p{10cm}lrrrr}
\toprule
Prompt & Model & $h$ & $\mathrm{H}$ & $\mathrm{H_{0.5}}$ & PPL \\
\cmidrule(){1-1} \cmidrule(l){2-2} \cmidrule(l){3-3} \cmidrule(l){4-4} \cmidrule(l){5-5} \cmidrule(l){6-6}
\multirow{5}{10cm}{Please complete the following sentence to make it as grammatically simple as possible} 
&  Llama-2 7B & 9.10 & 9.15 & 15.58 & 341.90 \\
&  Llama-2 13B & 9.21 & 8.37 & 15.18 & 184.71 \\
&  Llama-2 70B & 9.03 & 6.63 & 12.82 & 139.97 \\
&  Falcon 7B & 10.54 & 10.60 & 13.20 & 153.40 \\
&  Falcon 40B & 8.91 & 8.93 & 12.18 & 106.49 \\
&  GPT-3.5 D2 & 6.20 & - & - & 524.09 \\
&  GPT-3.5 D3 & 6.20 & - & - & 429.86 \\
\cmidrule(){1-1} \cmidrule(l){2-2} \cmidrule(l){3-3} \cmidrule(l){4-4} \cmidrule(l){5-5} \cmidrule(l){6-6}
\multirow{5}{10cm}{Please complete the following sentence with a careful focus on grammar} 
&  Llama-2 7B & 9.29 & 9.07 & 15.15 & 250.90 \\
&  Llama-2 13B & 9.36 & 7.85 & 14.51 & 169.02 \\
&  Llama-2 70B & 9.25 & 6.62 & 12.71 & 152.12 \\
&  Falcon 7B & 10.02 & 10.48 & 13.10 & 168.28 \\
&  Falcon 40B & 8.90 & 8.55 & 11.70 & 103.18 \\
&  GPT-3.5 D2 & 6.51 & - & - & 620.67 \\
&  GPT-3.5 D3 & 6.61 & - & - & 331.92 \\
\cmidrule(){1-1} \cmidrule(l){2-2} \cmidrule(l){3-3} \cmidrule(l){4-4} \cmidrule(l){5-5} \cmidrule(l){6-6}
\multirow{5}{10cm}{Please complete the following sentence to make it as grammatically complex as possible} 
&  Llama-2 7B & 9.12 & 8.28 & 14.30 & 315.23 \\
&  Llama-2 13B & 8.20 & 7.39 & 13.87 & 241.85 \\
&  Llama-2 70B & 8.35 & 5.73 & 12.02 & 191.69 \\
&  Falcon 7B & 10.25 & 10.39 & 12.98 & 143.58 \\
&  Falcon 40B & 8.67 & 8.80 & 12.15 & 105.21 \\
&  GPT-3.5 D2 & 5.81 & - & - & 1.014.37 \\
&  GPT-3.5 D3 & 6.00 & - & - & 458.79 \\
\cmidrule(){1-1} \cmidrule(l){2-2} \cmidrule(l){3-3} \cmidrule(l){4-4} \cmidrule(l){5-5} \cmidrule(l){6-6}
\multirow{5}{10cm}{Please complete the following sentence using the simplest vocabulary possible} 
&  Llama-2 7B & 9.01 & 9.27 & 15.45 & 362.27 \\
&  Llama-2 13B & 8.68 & 8.63 & 15.29 & 300.47 \\
&  Llama-2 70B & 8.60 & 6.52 & 13.07 & 214.54 \\
&  Falcon 7B & 10.41 & 10.77 & 13.09 & 167.87 \\
&  Falcon 40B & 8.57 & 8.72 & 12.16 & 115.59 \\
&  GPT-3.5 D2 & 5.89 & - & - & 638.71 \\
&  GPT-3.5 D3 & 5.96 & - & - & 630.27 \\
\cmidrule(){1-1} \cmidrule(l){2-2} \cmidrule(l){3-3} \cmidrule(l){4-4} \cmidrule(l){5-5} \cmidrule(l){6-6}
\multirow{5}{10cm}{Please complete the following sentence with a careful focus on word choice} 
&  Llama-2 7B & 9.10 & 8.58 & 14.72 & 269.79 \\
&  Llama-2 13B & 8.98 & 7.55 & 14.05 & 175.18 \\
&  Llama-2 70B & 8.80 & 6.42 & 12.51 & 168.91 \\
&  Falcon 7B & 9.64 & 10.06 & 12.90 & 166.03 \\
&  Falcon 40B & 8.67 & 8.30 & 11.65 & 105.26 \\
&  GPT-3.5 D2 & 6.73 & - & - & 453.51 \\
&  GPT-3.5 D3 & 6.53 & - & - & 329.08 \\
\cmidrule(){1-1} \cmidrule(l){2-2} \cmidrule(l){3-3} \cmidrule(l){4-4} \cmidrule(l){5-5} \cmidrule(l){6-6}
\multirow{5}{10cm}{Please complete the following sentence using the most difficult vocabulary possible} 
&  Llama-2 7B & 7.68 & 7.77 & 13.28 & 515.02 \\
&  Llama-2 13B & 7.23 & 7.42 & 13.34 & 376.41 \\
&  Llama-2 70B & 7.82 & 5.82 & 12.08 & 275.86 \\
&  Falcon 7B & 9.63 & 10.14 & 12.76 & 164.03 \\
&  Falcon 40B & 7.99 & 8.44 & 11.91 & 112.07 \\
&  GPT-3.5 D2 & 5.56 & - & - & 536.87 \\
&  GPT-3.5 D3 & 5.98 & - & - & 348.27 \\
\cmidrule(){1-1} \cmidrule(l){2-2} \cmidrule(l){3-3} \cmidrule(l){4-4} \cmidrule(l){5-5} \cmidrule(l){6-6}
\multirow{5}{10cm}{Please complete the following sentence in a human-like manner. It has been reported that human ability to predict next words is weaker than language models and that humans often make noisy predictions, such as careless grammatical errors} 
&  Llama-2 7B & 9.88 & 8.96 & 14.79 & 223.11 \\
&  Llama-2 13B & 9.85 & 8.40 & 14.70 & 170.97 \\
&  Llama-2 70B & 9.69 & 6.86 & 13.17 & 175.96 \\
&  Falcon 7B & 9.20 & 9.20 & 12.73 & 230.78 \\
&  Falcon 40B & 8.45 & 8.52 & 11.96 & 133.90 \\
&  GPT-3.5 D2 & 5.74 & - & - & 1.155.36 \\
&  GPT-3.5 D3 & 5.34 & - & - & 612.23 \\
\cmidrule(){1-1} \cmidrule(l){2-2} \cmidrule(l){3-3} \cmidrule(l){4-4} \cmidrule(l){5-5} \cmidrule(l){6-6}
\multirow{5}{10cm}{Please complete the following sentence. We are trying to reproduce human reading times with the word prediction probabilities you calculate, so please predict the next word like a human. It has been reported that human ability to predict next words is weaker than language models and that humans often make noisy predictions, such as careless grammatical errors} 
&  Llama-2 7B & 9.88 & 8.96 & 14.61 & 212.73 \\
&  Llama-2 13B & 9.38 & 8.43 & 14.44 & 192.83 \\
&  Llama-2 70B & 9.70 & 6.84 & 12.95 & 166.50 \\
&  Falcon 7B & 9.57 & 9.99 & 13.04 & 219.45 \\
&  Falcon 40B & 8.76 & 8.27 & 11.79 & 121.70 \\
&  GPT-3.5 D2 & 6.26 & - & - & 1.085.46 \\
&  GPT-3.5 D3 & 5.65 & - & - & 613.11 \\
\cmidrule(){1-1} \cmidrule(l){2-2} \cmidrule(l){3-3} \cmidrule(l){4-4} \cmidrule(l){5-5} \cmidrule(l){6-6}
\multirow{5}{10cm}{Please complete the following sentence} 
&  Llama-2 7B & 8.71 & 8.63 & 14.86 & 307.29 \\
&  Llama-2 13B & 8.74 & 8.16 & 14.52 & 191.69 \\
&  Llama-2 70B & 8.49 & 6.35 & 12.62 & 193.78 \\
&  Falcon 7B & 10.52 & 10.37 & 13.01 & 141.23 \\
&  Falcon 40B & 8.68 & 8.52 & 11.80 & 107.46 \\
&  GPT-3.5 D2 & 6.91 & - & - & 498.89 \\
&  GPT-3.5 D3 & 7.05 & - & - & 308.50 \\
\bottomrule
\end{tabular}
\caption{Full results of the experiments in \cref{sec:exp:prompt} on the DC with the format-1 prompts.}
\label{tbl:prompt_surprisal_dc}
\end{table*}

\begin{table*}[t]
\centering
\fontsize{8}{8}\selectfont
\setlength{\tabcolsep}{1.5pt}
\begin{tabular}{p{10cm}lrrrr}
\toprule
Prompt & Model & $h$ & $\mathrm{H}$ & $\mathrm{H_{0.5}}$ & PPL \\
\cmidrule(){1-1} \cmidrule(l){2-2} \cmidrule(l){3-3} \cmidrule(l){4-4} \cmidrule(l){5-5} \cmidrule(l){6-6}
\multirow{3}{10cm}{[INST] Please generate a grammatically simple sentence as much as possible. [/INST]  Answer:} 
&  LLaMA-2 7B & 8.88 & 7.07 & 13.72 & 325.44 \\
&  LLaMA-2 13B & 8.35 & 6.08 & 12.58 & 373.04 \\
&  LLaMA-2 70B & 7.38 & 4.67 & 10.59 & 466.80 \\
\cmidrule(){1-1} \cmidrule(l){2-2} \cmidrule(l){3-3} \cmidrule(l){4-4} \cmidrule(l){5-5} \cmidrule(l){6-6}
\multirow{3}{10cm}{[INST] Please generate a sentence with a careful focus on grammar. [/INST]  Answer:} 
&  LLaMA-2 7B & 9.25 & 6.79 & 13.48 & 232.50 \\
&  LLaMA-2 13B & 8.59 & 5.75 & 12.21 & 307.64 \\
&  LLaMA-2 70B & 8.05 & 4.15 & 10.13 & 345.80 \\
\cmidrule(){1-1} \cmidrule(l){2-2} \cmidrule(l){3-3} \cmidrule(l){4-4} \cmidrule(l){5-5} \cmidrule(l){6-6}
\multirow{3}{10cm}{[INST] Please generate a grammatically complex sentence as much as possible. [/INST]  Answer:} 
&  LLaMA-2 7B & 8.59 & 6.32 & 12.41 & 312.61 \\
&  LLaMA-2 13B & 8.35 & 5.68 & 11.94 & 329.54 \\
&  LLaMA-2 70B & 7.58 & 4.92 & 10.63 & 396.32 \\
\cmidrule(){1-1} \cmidrule(l){2-2} \cmidrule(l){3-3} \cmidrule(l){4-4} \cmidrule(l){5-5} \cmidrule(l){6-6}
\multirow{3}{10cm}{[INST] Please generate a sentence using the simplest vocabulary possible. [/INST]  Answer:} 
&  LLaMA-2 7B & 9.02 & 7.80 & 13.84 & 352.90 \\
&  LLaMA-2 13B & 7.55 & 6.34 & 12.75 & 564.19 \\
&  LLaMA-2 70B & 7.21 & 4.84 & 10.57 & 557.85 \\
\cmidrule(){1-1} \cmidrule(l){2-2} \cmidrule(l){3-3} \cmidrule(l){4-4} \cmidrule(l){5-5} \cmidrule(l){6-6}
\multirow{3}{10cm}{[INST] Please generate a sentence with a careful focus on word choice. [/INST]  Answer:} 
&  LLaMA-2 7B & 8.56 & 5.84 & 12.30 & 267.50 \\
&  LLaMA-2 13B & 8.00 & 5.63 & 11.85 & 327.15 \\
&  LLaMA-2 70B & 7.75 & 4.07 & 9.86 & 337.14 \\
\cmidrule(){1-1} \cmidrule(l){2-2} \cmidrule(l){3-3} \cmidrule(l){4-4} \cmidrule(l){5-5} \cmidrule(l){6-6}
\multirow{3}{10cm}{[INST] Please generate a sentence using the most difficult vocabulary possible. [/INST]  Answer:} 
&  LLaMA-2 7B & 8.22 & 5.93 & 12.05 & 351.65 \\
&  LLaMA-2 13B & 7.85 & 5.67 & 11.99 & 345.42 \\
&  LLaMA-2 70B & 7.28 & 4.43 & 10.29 & 597.32 \\
\cmidrule(){1-1} \cmidrule(l){2-2} \cmidrule(l){3-3} \cmidrule(l){4-4} \cmidrule(l){5-5} \cmidrule(l){6-6}
\multirow{3}{10cm}{[INST] Please generate a sentence in a human-like manner. It has been reported that human ability to predict next words is weaker than language models and that humans often make noisy predictions, such as careless grammatical errors. [/INST]  Answer:} 
&  LLaMA-2 7B & 8.97 & 7.19 & 13.46 & 317.99 \\
&  LLaMA-2 13B & 8.86 & 6.23 & 12.65 & 288.86 \\
&  LLaMA-2 70B & 8.65 & 4.56 & 11.05 & 364.23 \\
\cmidrule(){1-1} \cmidrule(l){2-2} \cmidrule(l){3-3} \cmidrule(l){4-4} \cmidrule(l){5-5} \cmidrule(l){6-6}
\multirow{5}{10cm}{[INST] Please generate a sentence. We are trying to reproduce human reading times with the word prediction probabilities you calculate, so please predict the next word like a human. It has been reported that human ability to predict next words is weaker than language models and that humans often make noisy predictions, such as careless grammatical errors. [/INST]  Answer:} 
&  LLaMA-2 7B & 9.15 & 7.95 & 14.33 & 292.45 \\
&  LLaMA-2 13B & 9.26 & 6.77 & 13.47 & 247.37 \\
&  LLaMA-2 70B & 8.30 & 4.86 & 11.52 & 318.45 \\
\\
\\
\cmidrule(){1-1} \cmidrule(l){2-2} \cmidrule(l){3-3} \cmidrule(l){4-4} \cmidrule(l){5-5} \cmidrule(l){6-6}
\multirow{3}{10cm}{[INST] Please generate a sentence. [/INST]  Answer:} 
&  LLaMA-2 7B & 9.08 & 6.61 & 13.34 & 223.67 \\
&  LLaMA-2 13B & 8.33 & 5.66 & 12.28 & 292.13 \\
&  LLaMA-2 70B & 7.85 & 4.68 & 10.84 & 298.61 \\
\bottomrule
\end{tabular}
\caption{Full results of the experiments in \cref{sec:exp:prompt} on the DC with the format-2 prompts.}
\label{tbl:prompt_surprisal_dc2}
\end{table*}

\begin{table*}[t]
\centering
\fontsize{8}{8}\selectfont
\setlength{\tabcolsep}{1.5pt}
\begin{tabular}{p{10cm}lrrrr}
\toprule
Prompt & Model & $h$ & $\mathrm{H}$ & $\mathrm{H_{0.5}}$ & PPL \\
 \cmidrule(){1-1} \cmidrule(l){2-2} \cmidrule(l){3-3} \cmidrule(l){4-4} \cmidrule(l){5-5} \cmidrule(l){6-6}
\multirow{5}{10cm}{Please complete the following sentence to make it as grammatically simple as possible} 
&  Llama-2 7B & 7.60 & 3.88 & 11.22 & 149.88 \\
&  Llama-2 13B & 6.42 & 3.04 & 10.68 & 84.31 \\
&  Llama-2 70B & 5.63 & 1.19 & 7.61 & 64.20 \\
&  Falcon 7B & 8.74 & 6.20 & 16.48 & 69.38 \\
&  Falcon 40B & 6.03 & 2.99 & 10.11 & 50.95 \\
&  GPT-3.5 D2 & 4.97 & - & - & 201.88 \\
&  GPT-3.5 D3 & 6.69 & - & - & 155.32 \\
\cmidrule(){1-1} \cmidrule(l){2-2} \cmidrule(l){3-3} \cmidrule(l){4-4} \cmidrule(l){5-5} \cmidrule(l){6-6}
\multirow{5}{10cm}{Please complete the following sentence with a careful focus on grammar} 
&  Llama-2 7B & 6.48 & 3.61 & 10.64 & 117.67 \\
&  Llama-2 13B & 5.32 & 3.20 & 10.39 & 84.06 \\
&  Llama-2 70B & 4.88 & 2.29 & 9.65 & 74.86 \\
&  Falcon 7B & 9.05 & 6.32 & 16.20 & 73.42 \\
&  Falcon 40B & 5.87 & 2.93 & 9.82 & 50.45 \\
&  GPT-3.5 D2 & 4.57 & - & - & 238.15 \\
&  GPT-3.5 D3 & 5.88 & - & - & 106.68 \\
\cmidrule(){1-1} \cmidrule(l){2-2} \cmidrule(l){3-3} \cmidrule(l){4-4} \cmidrule(l){5-5} \cmidrule(l){6-6}
\multirow{5}{10cm}{Please complete the following sentence to make it as grammatically complex as possible} 
&  Llama-2 7B & 6.69 & 4.14 & 10.32 & 118.23 \\
&  Llama-2 13B & 5.72 & 3.46 & 10.54 & 89.93 \\
&  Llama-2 70B & 4.79 & 1.54 & 7.73 & 75.42 \\
&  Falcon 7B & 8.51 & 5.36 & 15.12 & 62.39 \\
&  Falcon 40B & 5.69 & 2.82 & 9.58 & 46.98 \\
&  GPT-3.5 D2 & 3.64 & - & - & 353.98 \\
&  GPT-3.5 D3 & 4.49 & - & - & 149.71 \\
\cmidrule(){1-1} \cmidrule(l){2-2} \cmidrule(l){3-3} \cmidrule(l){4-4} \cmidrule(l){5-5} \cmidrule(l){6-6}
\multirow{5}{10cm}{Please complete the following sentence using the simplest vocabulary possible} 
&  Llama-2 7B & 6.74 & 4.18 & 11.88 & 158.02 \\
&  Llama-2 13B & 5.48 & 5.46 & 13.99 & 133.04 \\
&  Llama-2 70B & 4.83 & 2.93 & 10.67 & 94.71 \\
&  Falcon 7B & 7.92 & 6.51 & 16.49 & 69.94 \\
&  Falcon 40B & 5.62 & 2.97 & 10.03 & 53.52 \\
&  GPT-3.5 D2 & 4.66 & - & - & 245.26 \\
&  GPT-3.5 D3 & 4.92 & - & - & 214.47 \\
\cmidrule(){1-1} \cmidrule(l){2-2} \cmidrule(l){3-3} \cmidrule(l){4-4} \cmidrule(l){5-5} \cmidrule(l){6-6}
\multirow{5}{10cm}{Please complete the following sentence with a careful focus on word choice} 
&  Llama-2 7B & 5.98 & 3.50 & 1- & 127.13 \\
&  Llama-2 13B & 4.79 & 3.32 & 10.52 & 89.39 \\
&  Llama-2 70B & 4.74 & 2.71 & 10.29 & 82.48 \\
&  Falcon 7B & 8.78 & 5.53 & 15.58 & 71.02 \\
&  Falcon 40B & 5.82 & 2.89 & 9.93 & 50.66 \\
&  GPT-3.5 D2 & 4.92 & - & - & 174.62 \\
&  GPT-3.5 D3 & 5.67 & - & - & 107.04 \\
\cmidrule(){1-1} \cmidrule(l){2-2} \cmidrule(l){3-3} \cmidrule(l){4-4} \cmidrule(l){5-5} \cmidrule(l){6-6}
\multirow{5}{10cm}{Please complete the following sentence using the most difficult vocabulary possible} 
&  Llama-2 7B & 4.61 & 3.48 & 9.45 & 193.86 \\
&  Llama-2 13B & 4.44 & 3.26 & 9.83 & 148.45 \\
&  Llama-2 70B & 3.62 & 2.90 & 9.70 & 124.40 \\
&  Falcon 7B & 7.33 & 5.08 & 14.56 & 67.20 \\
&  Falcon 40B & 4.68 & 2.21 & 8.50 & 53.00 \\
&  GPT-3.5 D2 & 4.03 & - & - & 172.35 \\
&  GPT-3.5 D3 & 3.74 & - & - & 110.72 \\
\cmidrule(){1-1} \cmidrule(l){2-2} \cmidrule(l){3-3} \cmidrule(l){4-4} \cmidrule(l){5-5} \cmidrule(l){6-6}
\multirow{5}{10cm}{Please complete the following sentence in a human-like manner. It has been reported that human ability to predict next words is weaker than language models and that humans often make noisy predictions, such as careless grammatical errors} 
&  Llama-2 7B & 5.92 & 3.53 & 10.15 & 102.16 \\
&  Llama-2 13B & 5.16 & 3.64 & 10.76 & 80.88 \\
&  Llama-2 70B & 4.44 & 3.01 & 11.42 & 88.02 \\
&  Falcon 7B & 6.26 & 5.27 & 15.28 & 90.87 \\
&  Falcon 40B & 4.66 & 4.19 & 11.56 & 60.71 \\
&  GPT-3.5 D2 & 3.21 & - & - & 311.83 \\
&  GPT-3.5 D3 & 2.68 & - & - & 168.23 \\
\cmidrule(){1-1} \cmidrule(l){2-2} \cmidrule(l){3-3} \cmidrule(l){4-4} \cmidrule(l){5-5} \cmidrule(l){6-6}
\multirow{5}{10cm}{Please complete the following sentence. We are trying to reproduce human reading times with the word prediction probabilities you calculate, so please predict the next word like a human. It has been reported that human ability to predict next words is weaker than language models and that humans often make noisy predictions, such as careless grammatical errors} 
&  Llama-2 7B & 6.10 & 3.60 & 10.11 & 95.46 \\
&  Llama-2 13B & 4.95 & 3.83 & 10.83 & 81.96 \\
&  Llama-2 70B & 4.42 & 3.19 & 10.90 & 82.45 \\
&  Falcon 7B & 6.98 & 6.40 & 15.73 & 83.36 \\
&  Falcon 40B & 5.01 & 3.87 & 11.35 & 53.71 \\
&  GPT-3.5 D2 & 3.49 & - & - & 304.80 \\
&  GPT-3.5 D3 & 2.97 & - & - & 173.41 \\
\cmidrule(){1-1} \cmidrule(l){2-2} \cmidrule(l){3-3} \cmidrule(l){4-4} \cmidrule(l){5-5} \cmidrule(l){6-6}
\multirow{5}{10cm}{Please complete the following sentence} 
&  Llama-2 7B & 6.48 & 3.84 & 10.90 & 146.86 \\
&  Llama-2 13B & 5.32 & 3.60 & 11.34 & 95.65 \\
&  Llama-2 70B & 4.33 & 3.34 & 11.09 & 94.19 \\
&  Falcon 7B & 8.39 & 6.42 & 16.09 & 63.69 \\
&  Falcon 40B & 5.86 & 3.60 & 10.96 & 52.01 \\
&  GPT-3.5 D2 & 4.40 & - & - & 213.67 \\
&  GPT-3.5 D3 & 4.97 & - & - & 105.16 \\
\bottomrule
\end{tabular}
\caption{Full results of the experiments in \cref{sec:exp:prompt} on the NS with the format-1 prompts.}
\label{tbl:prompt_surprisal_ns}
\end{table*}

\begin{table*}[t]
\centering
\fontsize{8}{8}\selectfont
\setlength{\tabcolsep}{1.5pt}
\begin{tabular}{p{10cm}lrrrr}
\toprule
Prompt & Model & $h$ & $\mathrm{H}$ & $\mathrm{H_{0.5}}$ & PPL \\
\cmidrule(){1-1} \cmidrule(l){2-2} \cmidrule(l){3-3} \cmidrule(l){4-4} \cmidrule(l){5-5} \cmidrule(l){6-6}
\multirow{3}{10cm}{[INST] Please generate a grammatically simple sentence as much as possible. [/INST]  Answer:} 
&  LLaMA-2 7B & 4.60 & 3.04 & 9.35 & 146.17 \\
&  LLaMA-2 13B & 3.70 & 3.10 & 10.44 & 186.45 \\
&  LLaMA-2 70B & 3.22 & 1.57 & 6.55 & 193.25 \\
\cmidrule(){1-1} \cmidrule(l){2-2} \cmidrule(l){3-3} \cmidrule(l){4-4} \cmidrule(l){5-5} \cmidrule(l){6-6}
\multirow{3}{10cm}{[INST] Please generate a sentence with a careful focus on grammar. [/INST]  Answer:} 
&  LLaMA-2 7B & 5.04 & 2.59 & 8.44 & 100.95 \\
&  LLaMA-2 13B & 3.88 & 2.19 & 7.92 & 128.30 \\
&  LLaMA-2 70B & 4.09 & 0.87 & 5.06 & 140.79 \\
\cmidrule(){1-1} \cmidrule(l){2-2} \cmidrule(l){3-3} \cmidrule(l){4-4} \cmidrule(l){5-5} \cmidrule(l){6-6}
\multirow{3}{10cm}{[INST] Please generate a grammatically complex sentence as much as possible. [/INST]  Answer:} 
&  LLaMA-2 7B & 4.95 & 1.90 & 6.97 & 113.33 \\
&  LLaMA-2 13B & 4.72 & 1.28 & 7.07 & 115.80 \\
&  LLaMA-2 70B & 4.23 & 0.75 & 5.10 & 130.95 \\
\cmidrule(){1-1} \cmidrule(l){2-2} \cmidrule(l){3-3} \cmidrule(l){4-4} \cmidrule(l){5-5} \cmidrule(l){6-6}
\multirow{3}{10cm}{[INST] Please generate a sentence using the simplest vocabulary possible. [/INST]  Answer:} 
&  LLaMA-2 7B & 4.50 & 2.06 & 9.19 & 166.33 \\
&  LLaMA-2 13B & 3.39 & 2.32 & 9.64 & 270.99 \\
&  LLaMA-2 70B & 2.95 & 1.63 & 6.19 & 233.22 \\
\cmidrule(){1-1} \cmidrule(l){2-2} \cmidrule(l){3-3} \cmidrule(l){4-4} \cmidrule(l){5-5} \cmidrule(l){6-6}
\multirow{3}{10cm}{[INST] Please generate a sentence with a careful focus on word choice. [/INST]  Answer:} 
&  LLaMA-2 7B & 4.04 & 1.91 & 7.29 & 115.81 \\
&  LLaMA-2 13B & 3.75 & 1.52 & 6.74 & 146.38 \\
&  LLaMA-2 70B & 3.59 & 0.68 & 5.04 & 138.90 \\
\cmidrule(){1-1} \cmidrule(l){2-2} \cmidrule(l){3-3} \cmidrule(l){4-4} \cmidrule(l){5-5} \cmidrule(l){6-6}
\multirow{3}{10cm}{[INST] Please generate a sentence using the most difficult vocabulary possible. [/INST]  Answer:} 
&  LLaMA-2 7B & 4.32 & 1.21 & 6.25 & 151.10 \\
&  LLaMA-2 13B & 4.04 & 1.00 & 6.02 & 143.37 \\
&  LLaMA-2 70B & 2.62 & 1.14 & 5.74 & 221.19 \\
\cmidrule(){1-1} \cmidrule(l){2-2} \cmidrule(l){3-3} \cmidrule(l){4-4} \cmidrule(l){5-5} \cmidrule(l){6-6}
\multirow{3}{10cm}{[INST] Please generate a sentence in a human-like manner. It has been reported that human ability to predict next words is weaker than language models and that humans often make noisy predictions, such as careless grammatical errors. [/INST]  Answer:} 
&  LLaMA-2 7B & 4.24 & 3.54 & 10.99 & 125.88 \\
&  LLaMA-2 13B & 4.57 & 2.90 & 10.93 & 117.23 \\
&  LLaMA-2 70B & 4.08 & 1.38 & 7.00 & 143.37 \\
\cmidrule(){1-1} \cmidrule(l){2-2} \cmidrule(l){3-3} \cmidrule(l){4-4} \cmidrule(l){5-5} \cmidrule(l){6-6}
\multirow{5}{10cm}{[INST] Please generate a sentence. We are trying to reproduce human reading times with the word prediction probabilities you calculate, so please predict the next word like a human. It has been reported that human ability to predict next words is weaker than language models and that humans often make noisy predictions, such as careless grammatical errors. [/INST]  Answer:} 
&  LLaMA-2 7B & 5.22 & 3.34 & 10.49 & 118.79 \\
&  LLaMA-2 13B & 4.28 & 3.11 & 10.63 & 106.47 \\
&  LLaMA-2 70B & 4.79 & 1.46 & 7.17 & 127.61 \\
\\
\\
\cmidrule(){1-1} \cmidrule(l){2-2} \cmidrule(l){3-3} \cmidrule(l){4-4} \cmidrule(l){5-5} \cmidrule(l){6-6}
\multirow{3}{10cm}{[INST] Please generate a sentence. [/INST]  Answer:} 
&  LLaMA-2 7B & 5.38 & 2.98 & 8.91 & 91.26 \\
&  LLaMA-2 13B & 5.10 & 1.73 & 8.74 & 115.99 \\
&  LLaMA-2 70B & 4.33 & 1.12 & 5.82 & 121.74 \\
\bottomrule
\end{tabular}

\caption{Full results of the experiments in \cref{sec:exp:prompt} on the NS with the format-2 prompts.}
\label{tbl:prompt_surprisal_ns2}
\end{table*}

\section{Metalinguistic prompting}
\label{app:prompt_examples}
Tables~\ref{tbl:prompt_example_load} and~\ref{tbl:prompt_example_surprisal} show the exact prompts for directly asking LMs about the word's cognitive load or probability.
Note that a token ID is attached with respect to the token position in a sentence to distinguish the multiple tokens sharing the same word in a sentence.

\begin{table*}[t]
    \centering
        \fontsize{8}{9.5}\selectfont
\setlength{\tabcolsep}{1.5pt}
    \begin{tabular}{p{15cm}}
    \toprule
     Suppose humans read the following sentence: "'No, it's fine. I love it,' said Lucy knowing that affording the phone had been no small thing for her mother." \\
List the tokens and their IDs in order of their reading cost (high to low) during sentence processing. \\
Token ID: \\
0: 'No,, 1: it's, 2: fine., 3: I, 4: love, 5: it,', 6: said, 7: Lucy, 8: knowing, 9: that, 10: affording, 11: the, 12: phone, 13: had, 14: been, 15: no, 16: small, 17: thing, 18: for, 19: her, 20: mother., \\
Answer: \\
20: mother., 10: affording, 6: said, 11: the, 0: 'No,, 7: Lucy, 1: it's, 9: that, 17: thing, 5: it,', 2: fine., 15: no, 14: been, 3: I, 13: had, 8: knowing, 12: phone, 19: her, 16: small, 4: love, 18: for, \\
\\
Suppose humans read the following sentence: "A clear and joyous day it was and out on the wide open sea, thousands upon thousands of sparkling water drops, excited by getting to play in the ocean, danced all around." \\
List the tokens and their IDs in order of their reading cost (high to low) during sentence processing.\\
Token ID:\\
0: A, 1: clear, 2: and, 3: joyous, 4: day, 5: it, 6: was, 7: and, 8: out, 9: on, 10: the, 11: wide, 12: open, 13: sea,, 14: thousands, 15: upon, 16: thousands, 17: of, 18: sparkling, 19: water, 20: drops,, 21: excited, 22: by, 23: getting, 24: to, 25: play, 26: in, 27: the, 28: ocean,, 29: danced, 30: all, 31: around., \\
Answer:\\
13: sea,, 20: drops,, 28: ocean,, 21: excited, 0: A, 2: and, 22: by, 12: open, 7: and, 31: around., 19: water, 27: the, 3: joyous, 29: danced, 6: was, 23: getting, 11: wide, 18: sparkling, 30: all, 17: of, 14: thousands, 24: to, 15: upon, 4: day, 25: play, 1: clear, 8: out, 16: thousands, 5: it, 26: in, 9: on, 10: the,\\
\\
Suppose humans read the following sentence: "By the handsome reward many felt tempted, but the thought of the boar with its deadly tusks and face like thunder soon put an end to their ambitions." \\
List the tokens and their IDs in order of their reading cost (high to low) during sentence processing.\\
Token ID:\\
0: By, 1: the, 2: handsome, 3: reward, 4: many, 5: felt, 6: tempted,, 7: but, 8: the, 9: thought, 10: of, 11: the, 12: boar, 13: with, 14: its, 15: deadly, 16: tusks, 17: and, 18: face, 19: like, 20: thunder, 21: soon, 22: put, 23: an, 24: end, 25: to, 26: their, 27: ambitions., \\
Answer:\\
4: many, 27: ambitions., 3: reward, 5: felt, 2: handsome, 8: the, 6: tempted,, 26: their, 7: but, 21: soon, 1: the, 16: tusks, 12: boar, 19: like, 20: thunder, 13: with, 17: and, 14: its, 15: deadly, 22: put, 0: By, 10: of, 11: the, 18: face, 9: thought, 23: an, 24: end, 25: to,\\
\\
Suppose humans read the following sentence: <TARGET SENT> \\
List the tokens and their IDs in order of their reading cost (high to low) during sentence processing.\\
Token ID:\\
<TOKENS FROM TARGET SENT> \\
Answer:\\
\bottomrule
    \end{tabular}
    \caption{An example of a prompt for asking about the processing cost of words.}
    \label{tbl:prompt_example_load}
\end{table*}

\begin{table*}[t]
    \centering
        \fontsize{8}{9.5}\selectfont
\setlength{\tabcolsep}{1.5pt}
    \begin{tabular}{p{15cm}}
    \toprule
Suppose you read the following sentence: "'No, it's fine. I love it,' said Lucy knowing that affording the phone had been no small thing for her mother." \\
List the tokens and their IDs in order of their probability in context (low to high).\\
Token ID:\\
0: 'No,, 1: it's, 2: fine., 3: I, 4: love, 5: it,', 6: said, 7: Lucy, 8: knowing, 9: that, 10: affording, 11: the, 12: phone, 13: had, 14: been, 15: no, 16: small, 17: thing, 18: for, 19: her, 20: mother., \\
Answer:\\
0: 'No,, 10: affording, 8: knowing, 12: phone, 4: love, 5: it,', 7: Lucy, 15: no, 13: had, 17: thing, 1: it's, 6: said, 2: fine., 20: mother., 11: the, 18: for, 16: small, 9: that, 19: her, 3: I, 14: been,\\
\\
Suppose you read the following sentence: "A clear and joyous day it was and out on the wide open sea, thousands upon thousands of sparkling water drops, excited by getting to play in the ocean, danced all around." \\
List the tokens and their IDs in order of their probability in context (low to high).\\
Token ID:\\
0: A, 1: clear, 2: and, 3: joyous, 4: day, 5: it, 6: was, 7: and, 8: out, 9: on, 10: the, 11: wide, 12: open, 13: sea,, 14: thousands, 15: upon, 16: thousands, 17: of, 18: sparkling, 19: water, 20: drops,, 21: excited, 22: by, 23: getting, 24: to, 25: play, 26: in, 27: the, 28: ocean,, 29: danced, 30: all, 31: around., \\
Answer:\\
3: joyous, 21: excited, 14: thousands, 23: getting, 0: A, 8: out, 18: sparkling, 20: drops,, 1: clear, 5: it, 11: wide, 19: water, 30: all, 7: and, 15: upon, 28: ocean,, 29: danced, 13: sea,, 4: day, 9: on, 25: play, 31: around., 22: by, 24: to, 12: open, 2: and, 26: in, 6: was, 27: the, 10: the, 17: of, 16: thousands,\\
\\
Suppose you read the following sentence: "By the handsome reward many felt tempted, but the thought of the boar with its deadly tusks and face like thunder soon put an end to their ambitions." \\
List the tokens and their IDs in order of their probability in context (low to high).\\
Token ID:\\
0: By, 1: the, 2: handsome, 3: reward, 4: many, 5: felt, 6: tempted,, 7: but, 8: the, 9: thought, 10: of, 11: the, 12: boar, 13: with, 14: its, 15: deadly, 16: tusks, 17: and, 18: face, 19: like, 20: thunder, 21: soon, 22: put, 23: an, 24: end, 25: to, 26: their, 27: ambitions., \\
Answer:\\
2: handsome, 3: reward, 12: boar, 4: many, 18: face, 5: felt, 0: By, 6: tempted,, 21: soon, 9: thought, 20: thunder, 13: with, 15: deadly, 27: ambitions., 23: an, 7: but, 19: like, 1: the, 8: the, 17: and, 26: their, 11: the, 14: its, 22: put, 16: tusks, 10: of, 24: end, 25: to,\\
\\
Suppose you read the following sentence: <TARGET SENT> \\
List the tokens and their IDs in order of their probability in context (low to high).\\
Token ID:\\
<TOKENS FROM TARGET SENT> \\
Answer:\\
\bottomrule
    \end{tabular}
    \caption{An example of a prompt for asking about the word probability.}
    \label{tbl:prompt_example_surprisal}
\end{table*}

\end{document}